\newcommand{\system}{HapticBots}

\newcommand{\eyal}[1]{\textcolor{red}{#1}}
\newcommand{\change}[1]{\textcolor{red}{#1}}

\renewcommand{\eyal}[1]{#1}
\renewcommand{\change}[1]{#1}

\newcommand{\diff}[1]{\textcolor{purple}{#1}}
\renewcommand{\diff}[1]{#1}

\documentclass[sigconf]{acmart}

\AtBeginDocument{%
  \providecommand\BibTeX{{%
    \normalfont B\kern-0.5em{\scshape i\kern-0.25em b}\kern-0.8em\TeX}}}

\copyrightyear{2021}
\acmYear{2021}
\setcopyright{acmlicensed}\acmConference[UIST '21]{The 34th Annual ACM Symposium on User Interface Software and Technology}{October 10--14, 2021}{Virtual Event, USA}
\acmBooktitle{The 34th Annual ACM Symposium on User Interface Software and Technology (UIST '21), October 10--14, 2021, Virtual Event, USA} \acmPrice{15.00}
\acmDOI{10.1145/3472749.3474821} \acmISBN{978-1-4503-8635-7/21/10}

\begin{document}

\title{\system{}: Distributed Encountered-type Haptics for VR with Multiple Shape-changing Mobile Robots}

\author{Ryo Suzuki}
\affiliation{%
  \institution{University of Calgary}
  \city{Calgary, AB}
  \country{Canada}}
\email{ryo.suzuki@ucalgary.ca}

\author{Eyal Ofek}
\affiliation{%
  \institution{Microsoft Research}
  \city{Redmond, WA}
  \country{USA}}
\email{eyalofek@microsoft.com}

\author{Mike Sinclair}
\affiliation{%
  \institution{Microsoft Research}
  \city{Redmond, WA}
  \country{USA}}
\email{sinclair@microsoft.com}

\author{Daniel Leithinger}
\affiliation{%
  \institution{University of Colorado Boulder}
  \city{Boulder, CO}
  \country{USA}}
\email{daniel.leithinger@colorado.edu}

\author{Mar Gonzalez-Franco}
\affiliation{%
  \institution{Microsoft Research}
  \city{Redmond, WA}
  \country{USA}}
\email{margon@microsoft.com}

\renewcommand{\shortauthors}{Suzuki, et al.}

\begin{abstract}
{\em HapticBots} introduces a novel encountered-type haptic approach for Virtual Reality (VR) based on multiple tabletop-size shape-changing robots. These robots move on a tabletop and change their height and orientation to haptically render various surfaces and objects on-demand. Compared to previous encountered-type haptic approaches like shape displays or robotic arms, our proposed approach has an advantage in deployability, scalability, and generalizability---these robots can be easily deployed due to their compact form factor. They can support multiple concurrent touch points in a large area thanks to the distributed nature of the robots. We propose and evaluate a novel set of interactions enabled by these robots which include: 1) rendering haptics for VR objects by providing just-in-time touch-points on the user's hand, 2) simulating continuous surfaces with the concurrent height and position change, and 3) enabling the user to pick up and move VR objects through graspable proxy objects. Finally, we demonstrate HapticBots with various applications, including remote collaboration, education and training, design and 3D modeling, and gaming and entertainment.
\end{abstract}




\begin{teaserfigure}
\centering
\includegraphics[width=1\textwidth]{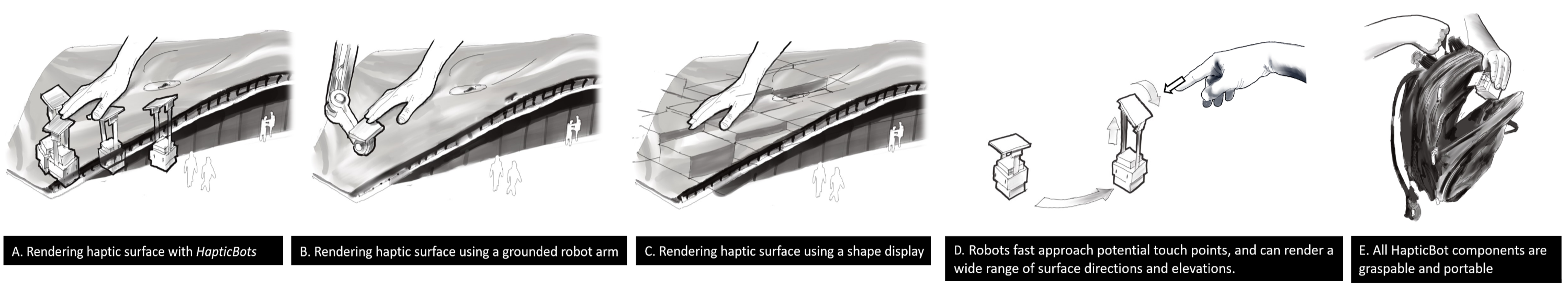}
\caption{(a) {\em HapticBots} enable to support multiple concurrent touch points of a virtual surface, giving the illusion of a large physical object. An existing approach such as using a grounded robot arm (b) to bring a touchable surface piece to encounter the user hand ~\cite{araujo2016snake,vonach2017vrrobot} is limited to render a single touch point. (c) shape displays ~\cite{follmer2013inform, leithinger2015shape} may render the entire shape simultaneously but are limited in fidelity and the area they can cover (display geometry resolution is displayed courser for visualization).
(d) The HapticBots are designed to coordinate and move fast to encounter the hand when it reaches possible touch points and render 5 degrees of freedom (3D position \& 2 angles of surface normal). (e) all the system's components are light, portable, and easy to deploy.}
\Description{Five sketches are depicted to illustrate the HapticBots' features. a) A virtual architecture model in VR and the user's hand on four HapticBots are depicted to show how the HapticBots can render the haptic touchpoints of the virtual model. b) The same virtual architecture model in VR and the user's hand on a robotic arm are depicted. c) The same virtual architecture model in VR and the user's hand on a shape display are depicted. d) One HapticBots approaches to a user's hand to provide a touchpoint by moving, changing its height, and tilting. e) The user is bringing out HapticBots from her backpack.}
\label{fig:teaser}
\end{teaserfigure}

\maketitle

\begin{figure*}[t!]
\centering
\includegraphics[width=1\textwidth]{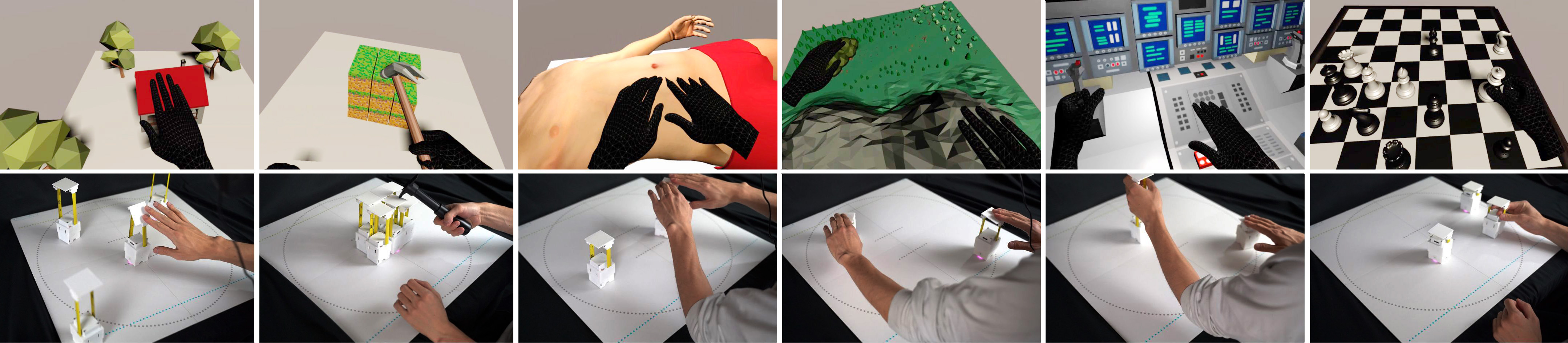}
\caption{HapticBots render encountered-type haptics for a range of VR applications.}
\Description{
Five different applications for HapticBots shown in 10 subfigures arranged in two rows.
Top row from left to right:
First panel: A virtual reality view of a house scene with trees on a tabletop, with the user's hand touching the house.
Second panel: A virtual reality view of a game scene with blocky terrain on a tabletop, with the hammering on a block.
Third panel: A virtual person's body, with the users hands palpating the stomach.
Fourth panel: A virtual landscape with the user touching a part of it.
Fifth panel: A virtual spaceship control panel, with the user touching a button and a joystick.
The bottom row depicts the physical HapticBots system configuration that corresponds to the application.
First panel: Three robots on a tabletop with the user’s hand touching the center robot.
Second panel: Four robots on a tabletop with the user hammering on one of the robot with a hammer.
Third panel: Three robots on a tabletop with the user’s two hands touching the leftmost robot.
Fourth panel: Two robots on a tabletop with the user’s two hands touching the two robots.
Fifth panel depicts two robots on a tabletop with the user’s two hands touching the two robots.
}
\label{fig:apps}
\end{figure*}

\section{Introduction}
Effective haptic feedback promises to enrich Virtual Reality (VR) experiences in many application  domains~\cite{hayward2004haptic}, 
\change{but supporting general-purpose haptic feedback is still a difficult challenge.
A common approach to providing haptic feedback for VR is to use a hand-held or wearable device~\cite{choi2016wolverine, choi2018claw, heo2018thor, Kovacs2020HapticPivot, lee2019torc}.
However, these wearable hand-grounded devices are inherently limited in their ability to render a world grounded force, such as  surfaces that can be touched or pushed with the user's hand.}

To fill this gap, encountered-type haptics~\cite{yokokohji2005designing, mcneely1993robotic} are introduced as an alternative approach.
\change{In contrast to hand-held or wearable devices, the encountered-type haptics provide haptic sensations through {\it actuated physical environments} by dynamically moving physical objects~\cite{he2017physhare, suzuki2020roomshift, araujo2016snake} or transforming the physical shape~\cite{siu2018shapeshift, nakagaki2019inforce} when the user encounters the virtual object.}

Different approaches have been developed for encountered-type haptics: from grounded robotic arms (e.g., Snake Charmer~\cite{araujo2016snake}, VRRobot~\cite{vonach2017vrrobot}) to 
shape displays (e.g., shapeShift~\cite{siu2018shapeshift}, Feelex~\cite{iwata2001project}, inForce~\cite{nakagaki2019inforce}). 
\change{However, the current approaches still face a number of challenges and limitations.}
For example, shape displays (Figure \ref{fig:teaser} c)
\change{often require  large, heavy, and mechanically complex devices, reducing reliability and deployability of the system for use outside research labs.}
Also, the resolution fidelity and the display's size are still limited, making it difficult to render smooth and continuous surfaces across a large interaction area. 
Alternately, robotic arms (Figure \ref{fig:teaser} b) can bring a small piece of a surface to meet the user hand on demand, but the speed at which humans move challenges the ability to cover just in time large interaction spaces with a single device.
\change{
Scaling the number of robotic arms is also a challenge as complex 3D path planning is required to avoid unnecessary collision with both the user and the other arms.
}

\change{
The goal of this paper is to address these challenges by introducing a novel encountered-type haptics approach, which we call {\it distributed encountered-type haptics} (Figure~\ref{fig:teaser}).
Distributed encountered-type haptics employ multiple shape-changing mobile robots to simulate a consistent physical object that the user can encounter through hands or fingers.
By synchronously controlling multiple robots, these robots can approximate different objects and surfaces distributed in a large interaction area.
}

\change{
Our proposed approach enables {\it deployable}, {\it scalable}, and {\it general-purpose} encountered-type haptics for VR, providing a number of advantages compared to the existing approaches, including shape displays~\cite{siu2018shapeshift, iwata2001project,abtahi2018visuo,nakagaki2019inforce}, robotic arms~\cite{araujo2016snake, vonach2017vrrobot}, and non-transformable mobile robots~\cite{he2017physhare, gonzalez2020reach}.
{\bf 1) Deployable:} Each mobile robot is light and compact, making the system portable and easy to deploy (Figure \ref{fig:teaser} e).
{\bf 2) Scalable:} Since each robot is simple and modular, it can scale to increase the number of touch-points and covered area.
Moreover, the use of multiple robots can reduce the average distance that a robot needs to travel, which reduces the robots' speed requirements.
{\bf 3) General-purpose:} Finally, the shape-changing capability of each robot can significantly increase the expressiveness of haptic rendering by transforming itself to closely match with the virtual object on-demand and in real-time.
This allows for greater flexibility needed for general-purpose applications.
}

The main contribution of this paper is a concept of this novel encountered  haptic approach and a set of distributed interaction techniques, outlined in Section 3 (e.g., Figure~\ref{fig:concept}-\ref{fig:lateral-motion}).
To demonstrate this idea, we built {\em \system{}}, an open source~\footnote{\url{https://github.com/ryosuzuki/hapticbots}} tabletop shape-changing mobile robots that are specifically designed for distributed encountered-type haptics.
\system{} consists of off-the-shelf mobile robots (Sony TOIO), and custom height-changing mechanisms to haptically render general large surfaces with varying normal directions (-60 to 60 degrees).
It can cover a large space (55 cm $\times$ 55 cm) above the table (a dynamic range of 24 cm elevation) at high speed (24 cm/sec and 2.8 cm/sec for horizontal and vertical speed, respectively).
Each robot is compact (4.7 $\times$ 4.7 $\times$ 8 cm, 135 g) and its tracking system consists of an expandable, pattern-printed paper mat; thus, it is portable and deployable.

\change{
Our HapticBots' hardware design is inspired by ShapeBots~\cite{suzuki2019shapebots}, but as far as we know, our system is the first exploration of using multiple tabletop shape-changing robots for VR haptics. 
Applying to VR haptics introduces a set of challenging requirements, which led to a new distributed haptics system design as well as to new hardware for each of the robots:
{\it 1) Efficient path planning integrated with real-time hand tracking:}
The system coordinates the movements of all robots with the user's hand. We track and anticipate potential touch points at a high frame rate (60 FPS) and guide the robots to encounter the user's hands in a just in time fashion.
{\it 2) Precise height and tilt control:} In contrast to ShapeBots' open-loop system, \system{} enables more precise height and tilt control with embedded encoders and closed-loop control system to render surfaces with varying normal angles.
{\it 3) Actuator robustness:} We vastly improved actuator force by around 70x (21.8N vs. 0.3N holding force of ~\cite{suzuki2019shapebots}) to provide meaningful force feedback.
In addition to these technical contributions, we developed various VR applications to demonstrate the new possibilities for encountered haptics, including remote collaboration, medical training, 3D modeling, and entertainment.
}

To evaluate how \system{} can create plausible haptic sensations, we conducted two user studies. In the first task, participants were asked to encounter and explore the surfaces of different virtual objects and report the encounters' level of realism. In the second task, we measured the capabilities of \system{} to deliver continuous touch sensation when exploring  surfaces, and the level of granularity that it can provide by asking participants to distinguish among surfaces of different tilt angles. We further compared the haptic sensations between \system{} and a static shape display.
Our study validates the effectiveness of the proposed approach.

Finally, the contributions of this paper are: 
\begin{enumerate}
\item \diff{A novel concept of distributed encountered-type haptics}
\item \diff{A hardware and software implementation of \system{} and its applications}
\item \diff{User evaluations that investigate the effects of \system{} haptic illusions.}
\end{enumerate}

\section{Related Work}

\subsection{Haptic Interfaces for VR}

\subsubsection{Hand-held and Wearable Haptic Interfaces}
In recent years, various haptic devices have been explored to enhance the user immersion of VR.
One of the most common types of haptic devices is the hand-held or wearable haptic approach. 
\eyal{Most hand-held haptic devices render touch sensations of virtual objects by applying differential forces of the fingertips against the opposing or thumb~\cite{benko2016normaltouch, Kovacs2020HapticPivot,sun2019PoCoPo}.
For realizing the dynamic range of the device movement, the fingertip is usually pushed back to stay outside the virtual object until interaction~\cite{choi2018claw, Sinclair2019Capstan, choi2016wolverine}. 
Only a few devices, such as Haptic Revolver ~\cite{whitmire2018haptic}, can also render forces such as texture and cutaneous shear force of the virtual surface.} 
However, one inherent limitation of such {\it body grounded} devices is the lack of generating a convincing {\it world grounded} sensation as no perceived force stops the body from moving into virtual objects. 
Grounding the hands to the body using exoskeletons or strings~\cite{wireality2020} can aid the grounding perception but are cumbersome and complex. 

\subsubsection{Passive Haptics}
Alternatively, {\it passive haptics}~\cite{insko2001passive} approach uses a physical artifacts such as a haptic proxy for VR, so that a VR user can touch and interact with a real object.
For example, Annexing Reality~\cite{hettiarachchi2016annexing} employs static physical objects as props in an immersive environment by matching and adjusting the shape and size of the virtual objects.
Haptic Retargeting~\cite{azmandian2016haptic, cheng2017sparse} leverages a mismatch in hand-eye coordination in order to guide the user's touch toward the position of physical objects. 
Similarly, by combining passive objects with redirected walking~\cite{razzaque2005redirected}, Kohli et al.~\cite{kohli2005combining} explored haptics that can go beyond the scale of human hands. Using passive objects, one can generate very reasonable haptic sensations. \eyal{However, as the shape and position of the proxy object in this case is fixed, it has a limited degree of haptic expression. For example, when the position or geometry of the proxy object differs from the displayed virtual object, it can break the illusion~\cite{simeone2015substitutional}.}
Manual reconfiguration of proxy objects has been also explored~\cite{arora2019virtualbricks, zhu2019haptwist}, but lacks the capability of dynamically simulating various shapes on demand.

\subsubsection{Robotic Encountered-type Haptics}
To overcome this limitation, McNeely proposed {\it robotic encountered-type haptics} by integrating a passive haptic proxy with mechanical actuation~\cite{mcneely1993robotic}.
Encountered-type haptics dynamically positions or transform the haptic props when the user ``encounters'' the virtual object.
Overall, there are three different approaches that have been explored for tabletop encountered-type haptics: robotic arms, shape displays, and mobile robots.

Robotic arms~\cite{hirota1995simulation}, such as SnakeCharmer~\cite{araujo2016snake} and VRRobot~\cite{vonach2017vrrobot} simulate surfaces by bringing a small patch of a surface to the user's hand wherever she may touch the virtual surface.
Since the virtual object is invisible to the VR user,  it can potentially generate the perception that the entire geometry exists in the physical world. However, the need for a robot arm to cover a large interaction space requires a large arm with a long reach which may be heavy and less potable. Also, the requirement for moving the large robotic arm in a volume while the user is blind to it, may limit the speed or movement space of the robot for safety reasons.

The second approach is shape displays~\cite{follmer2013inform, leithinger2015shape, steed2021mechatronic}. 
Systems like Feelex~\cite{iwata2001project},
inFORCE~\cite{nakagaki2019inforce}, and shapeShift~\cite{abtahi2018visuo, siu2018shapeshift} simulate dynamic surfaces and shapes by constructing the encountered geometry using an array of actuated pins.
\eyal{However, the large number of actuators that are needed to render a shape limits these devices' resolution and makes them complex, expensive, heavy, power hungry, and limited in coverage area.}

The third approach uses mobile robots~\cite{he2017physhare, wang2020movevr, suzuki2020roomshift, yixian2020zoomwalls, gonzalez2020reach, iwata1999walking} or even drones~\cite{abtahi2019beyond, hoppe2018vrhapticdrones} to move or actuate passive proxy objects.
For example, PhyShare~\cite{he2017physhare} and REACH++~\cite{gonzalez2020reach} employ the tabletop mobile robots to dynamically reposition the attached passive haptic proxy.
However, these mobile robots can only render a single predefined object due to the lack of transformation capability.
\eyal{Zhao et al.~\cite{zhao2017robotic} explored assembled haptic proxies of objects using swarm robots. 
While they assemble the required geometry on demand, it requires significant time to assemble a large object, which limits real-time interaction.

As we can see, the existing encountered-type approaches still have many challenges in terms of deployability (portable and deployable form factor), scalability (both an interaction area and the number of touch-points), and generalizability (the variety of shapes and surfaces the system can support). 
Our contribution is to address these problems by introducing a new class of encountered-type haptics with distributed shape-changing robots. 
}

\begin{figure*}[!ht]
\centering
\includegraphics[width=1\textwidth]{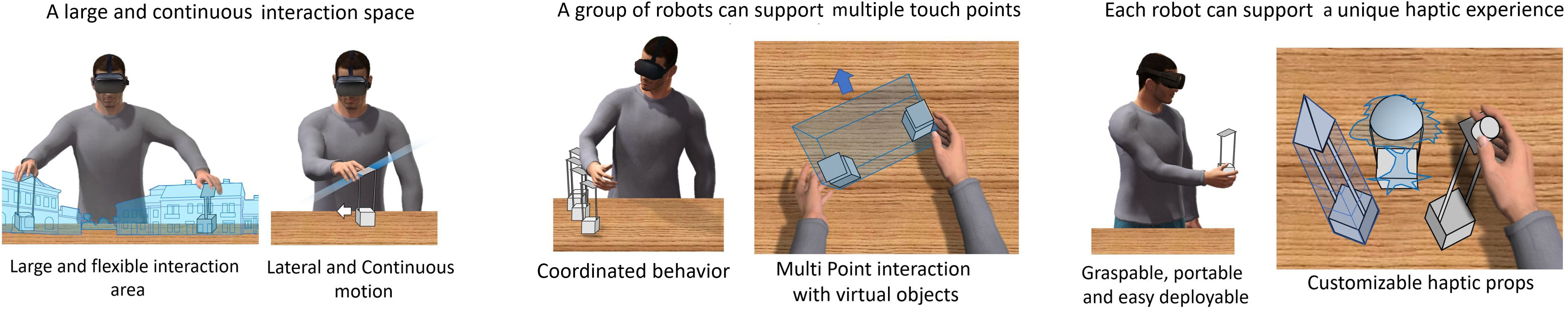}
\caption{Unique features of distributed encountered-type haptics.
\eyal{First, our freely moving robots can reach and render touch points in a large interaction space. Accurate rendering of 3D locations and orientations of surfaces (left). 
The use of multiple robots can generate multiple touch events concurrently (middle). Finally, each robot can be regarded as a separate object that may be picked up, moved around, and the payload that it carries can be uniquely fitted to the application}}
\Description{
The unique features of HapticBots shown in 6 subfigures arranged in two rows. From left to right: The first figure top: A person wearing a VR headset touching two robots on a tabletop, with a virtual city scene with buildings overlaid as a semi transparent graphic.
Second figure top: A person wearing a VR headset touching a moving robot on a tabletop. An arrow indicates the robots moving direction.
Third figure top: A person wearing a VR headset picking up a robot.
The first figure bottom: Top view of two hands touching two robots on a tabletop
Second figure bottom: A person wearing a VR headset resting their arm on four tabletop robots.
Third figure bottom: A hand touching one of three robots on a tabletop, with custom props attached to their top.
}
~\label{fig:concept}
\end{figure*}

\subsection{Swarm User Interfaces}
Our work is also built on recent advances in swarm user interfaces, which leverage a swarm of robots for tangible and shape-changing interactions (e.g., Zooids~\cite{le2016zooids}, UbiSwarm~\cite{kim2017ubiswarm},
HERMITS~\cite{nakagaki2020hermits}, ShapeBots~\cite{suzuki2019shapebots}, Reactile~\cite{suzuki2018reactile}, PICO~\cite{patten2007mechanical}).
Some of the previous systems have demonstrated the haptic and tactile sensation with swarm interfaces. For example, SwarmHaptics~\cite{kim2019swarmhaptics} demonstrate the use of swarm robots for everyday, non VR, haptic interactions (e.g., notification, communication, force feedback) and RoboGraphics~\cite{guinness2019robographics} and FluxMarker~\cite{suzuki2017fluxmarker} explores the tactile sensation for people with visual impairments. 
More recently, several works have been introduced to augment the capability of each robot to enhance interactions. 
For example, HERMITS~\cite{nakagaki2020hermits} augment the robots with customizable mechanical add-ons to expand tangible interactions.

Particularly, our work is inspired by the idea of {\it ``shape-changing swarm robots''} introduced by ShapeBots~\cite{suzuki2019shapebots}.
ShapeBots demonstrates the idea of combining a small table-top robot with a miniature reel actuator to greatly enhance the range of interactions and expressions for tangible user interfaces.

\change{
However, none of these works are aimed at rendering haptics of general large geometries in VR. 
As far as we know, our system is the first exploration of using tabletop-size shape-changing swarm robots for VR haptics.
Applying swarm UIs to VR haptics introduces a set of challenges and opportunities.
For example, the prior work~\cite{suzuki2019shapebots} explicitly articulated that support for AR/VR haptics is their limitation and future work due to a number of technical challenges, including the robustness of the actuation and VR software integration. 
On the other hand, in VR, the user is blind to the real world, thus it is possible to render larger geometries with only a smaller number of just in time robots.
Our goals are to explore this previously unexplored design space, introduce a set of haptic interactions, and address these software and hardware challenges for VR haptics applications.
}

\section{Distributed Encountered-type Haptics}
\change{
\subsection{Concept and Unique Features}
This section introduces a novel encountered-type haptic approach, which we call {\it distributed encountered-type haptics}.
Distributed encountered type haptics employ multiple coordinated robots that can move their position and transform their shape to haptically render various objects and surfaces distributed in the space.
Our approach has the following unique features: 1) support for large and flexible interaction areas, and 2) portable and deployable form factor.
}

\subsubsection{Large and Flexible Interaction Area}
One of the unique advantages of our approach is the ability for distributed and fast moving robots to cover a large and flexible interaction space (Figure~\ref{fig:concept}) and
 leverage two-handed interactions. 
Since each robot is simple and modular, it is easy to scale the number of touch points and covered area.

\subsubsection{Portable and Deployable Form Factor}
Distributed robots are composed of compact and small components, and are not bound to preset locations.
Our implementation particularly leverages recent advantages of tracking systems in both the VR headset and the robot's location.
For example, \system{} uses a lightweight mat, printed with a dot pattern viewed by the robots, as a tracking mechanism. Since the setup of this tracking mechanism is fairy simple (only placing a mat), the system can work in any flat location without dedicated external tracking devices (e.g., a transparent desk and camera in~\cite{suzuki2019shapebots} or optical tracking systems in~\cite{suzuki2020roomshift, he2017physhare, gonzalez2020reach}).
Also, a standalone Oculus Quest HMD enables inside-out hand tracking, where the system does not require any calibrated setup.
Since each robot is compact, it is possible to put the entire system in a small carrying case and quickly deploy it to another horizontal surface. 

\begin{figure}[h!]
\centering
\resizebox{\columnwidth}{!}{
\includegraphics[height=1\textwidth]{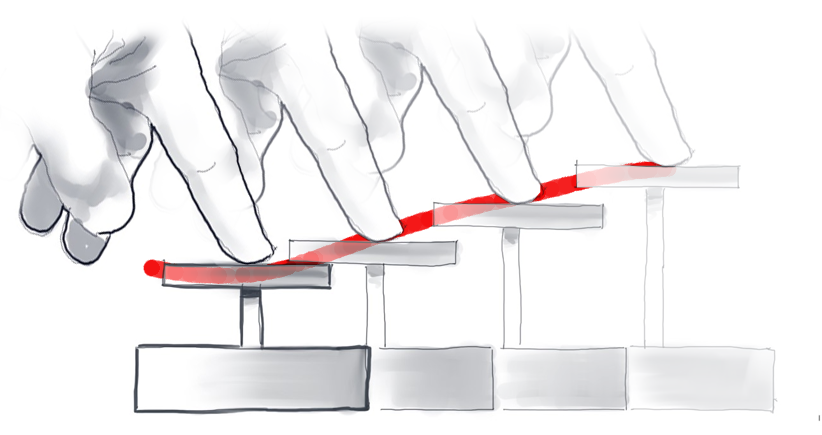}
\hspace{1cm}
\includegraphics[height=1\textwidth]{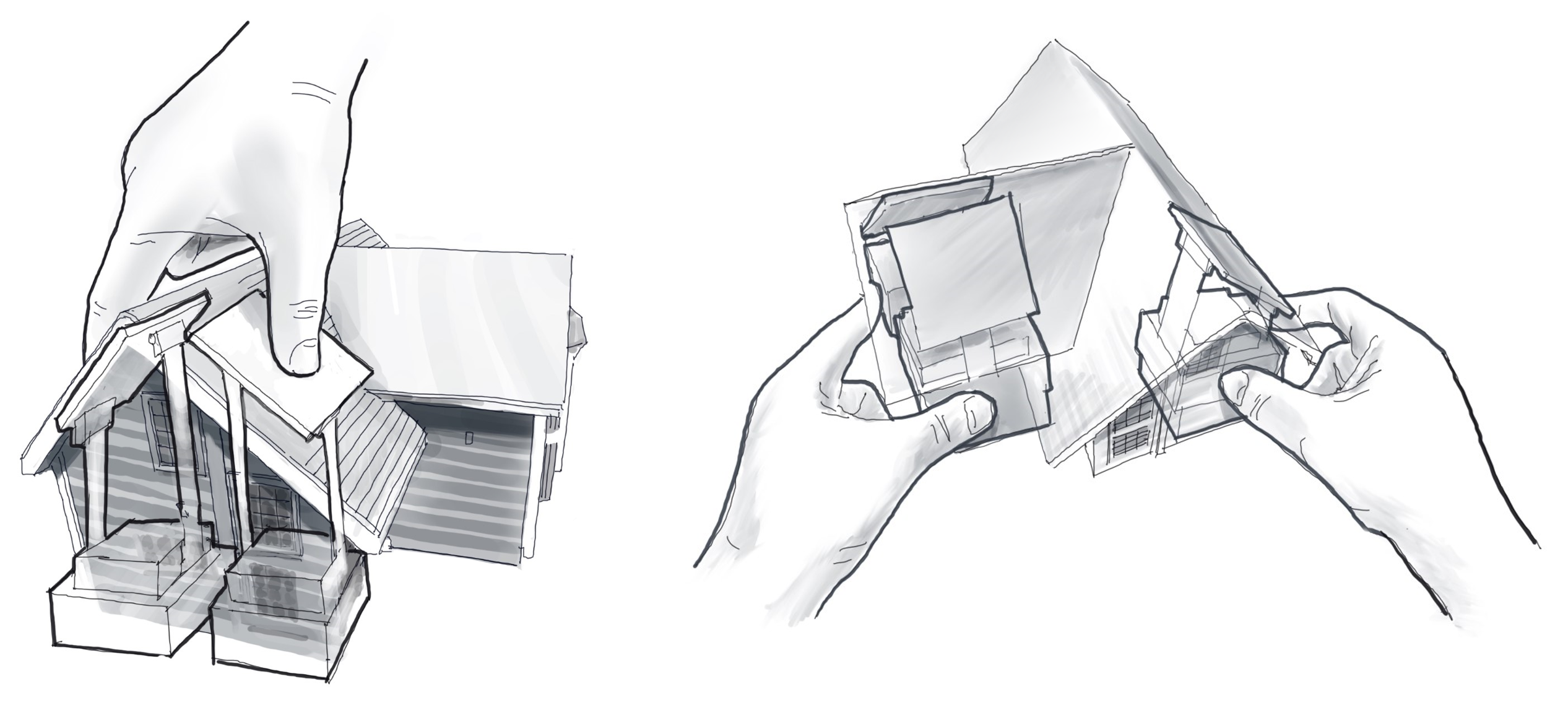}
}
\caption{Left: \system{} supports lateral movements of the touch surface (shown as a red curve). The robot may also tilt its surface to better reflect the surface normal. Right: Coordinated behaviors with multiple robots. 
Multiple touch points can enable the user to move, rotate or scale virtual objects.
\Description{
Left: HapticBots supports lateral movements of the touch surface. A finger rests on a moving robot, which extends to represent a slope as it moves.
Right: Coordinated behaviours with multiple robots, depicted left to right: Side view of a hand touching three robots with tilting tops. A finger touches a single robot with a tilting tops that represents a virtual house model. Two hands touch four robots at their side. Two hands feel a virtual house model by touching two robots
}
}
~\label{fig:lateral-motion}
\end{figure}

\subsection{Haptic Interactions and Unique Affordances}
In addition to these unique features, our approach enables the following haptic interactions and unique affordances. 

\subsubsection{
Rendering Continuous Surfaces with Concurrent Lateral Motion.}
Pin-based shape displays approximate a shape of a surface by a set of discreet pins, which generates spatial aliasing when rendering diagonal or curved objects~\cite{abtahi2018visuo}. Particularly when sliding ones fingers over a non-horizontal surface of the shape display, the abrupt shape change of the individual pins are very noticeable. 
\eyal{On the other hand, our approach can closely render the continuous smooth surface by leveraging the concurrent lateral motion. 
By using real-time tracking of the user's hands to guide the robots, it helps enable constant contact with the user's fingers. 
Robots can follow the movement of the user's finger and move or rotate in azimuth with the hand while changing its surface height and orientation to follow the virtual geometry (Figure~\ref{fig:lateral-motion}).}

\subsubsection{Rendering Large Objects with Coordinated Multiple Robots}
The coordination of multiple robots can extend their rendering capabilities. 
For example, Figure~\ref{fig:lateral-motion} illustrates an example of multi-point interaction with coordinated robots to simulate the haptic sensation of an object that is much larger than each robot.
\eyal{A group of robots can mimic a small shape display with additional degrees of freedom such as tilting and relative rotations. For example, two tilted robots can work in tandem to simulate a corner between two surfaces, such as the tip of a roof of a virtual house (Figure~\ref{fig:lateral-motion})}

\subsubsection{Rendering a Large Number of Objects with a Smaller Number of Robots}
These robots can also give the illusion that the user can touch more objects than the actual number of robots.
Even with a smaller number of robots, by leveraging the locomotion capability and anticipation of hand movement, the robots can position themselves to the object which the user will most likely encounter in the next moment. 
With that, the user may perceive that the entire scene is haptically rendered with these robots.

\subsubsection{Rendering Graspable Objects for Tangible Interaction}
Another unique affordance of the mobile robot is its graspability. (Figure~\ref{fig:concept}).
A compact robot can be picked and moved to another location by the user. This ability opens new interaction possibilities such as positioning robots as input (e.g. as proxies to virtual objects).

\begin{figure}[h!]
\centering
\resizebox{\columnwidth}{!}{
\includegraphics[height=1\textwidth]{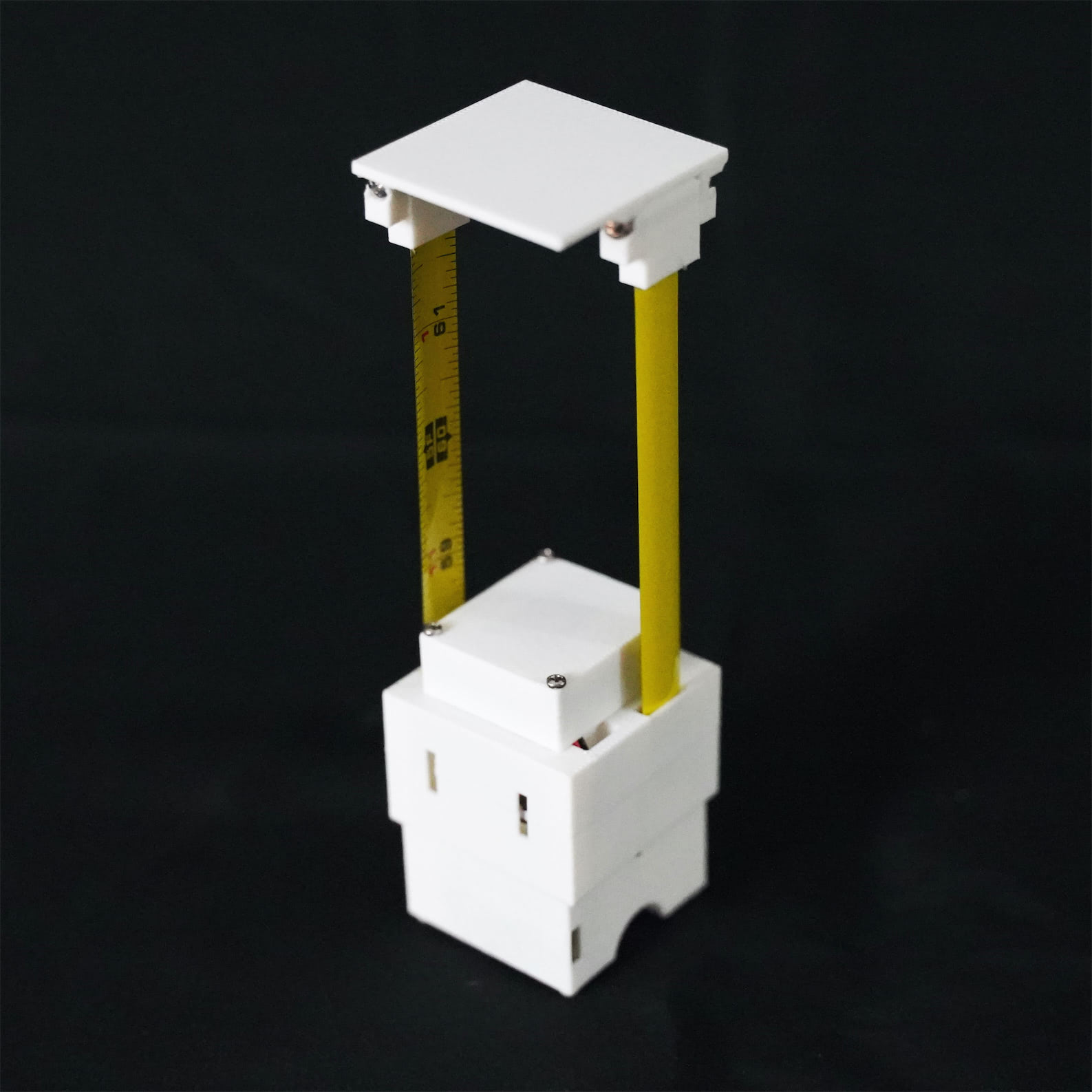}
\includegraphics[height=1\textwidth]{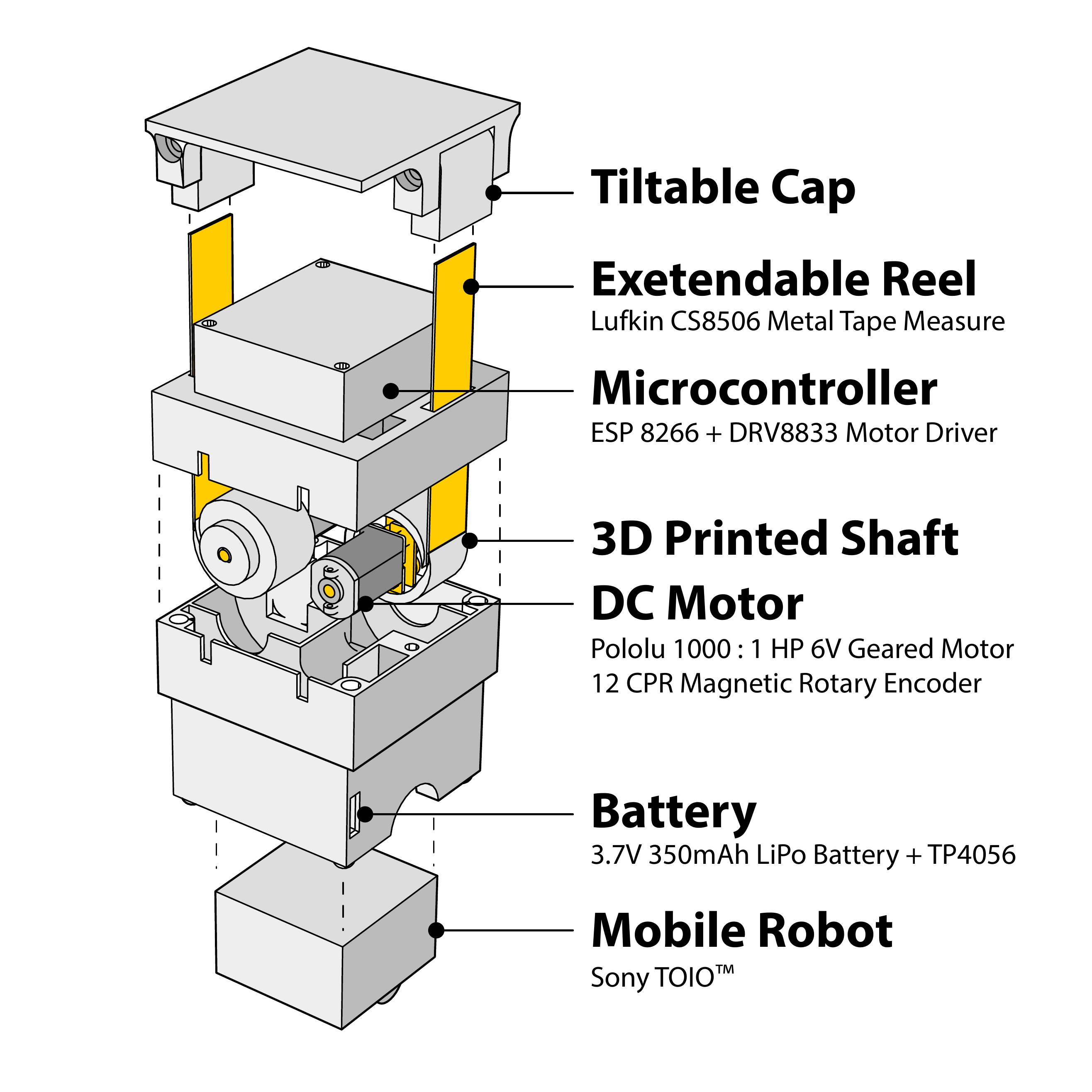}
}
\caption{Mechanical design of the reel actuator.}
\Description{
Mechanical Design of the reel actuator. A photo of a robot with an extended actuator. Next to it an exploded CAD view of the robot, showing the internal mechanics, consisting of Tiltable Cap, Extendable Reel, Microcontroller, 3D Printed Shaft DC Motor, Battery, and Mobile Robot Sony Toio
}
~\label{fig:mechanical-design}
\end{figure}

\section{\system{}: System and Implementation}

To demonstrate our concept, we built \system{}, a system that consists of multiple coordinated height-changing robots and the associated VR software.
Each robot is made of 1) custom-built shape-changing mechanisms with reel-based actuators, and 2) an off-the-shelf mobile robot (Sony Toio) that can move on a mat printed with a pattern for position tracking.
For the VR system, we used Oculus Quest HMD and its hand tracking capability for interaction.
The software system synchronizes virtual scenes with physical environment (e.g., each robot's position, orientation, and height), so that the robots can provide a haptic sensation in a timely manner. 
This section describes the design and implementation of the both hardware and software systems, then provides technical evaluation of \system{} prototype.

\subsection{Reel-based Linear Actuator}
\subsubsection{Mechanical Design}
To enable a large dynamic range of heights with a compact form factor, \system{} employs an extendable reel-based linear actuator, inspired by prior works (e.g., ShapeBots~\cite{suzuki2019shapebots}, KineReels~\cite{takei2011kinereels}, and G-Raff~\cite{kim2015g}). 
However, we need to accommodate for: 1) mechanical stability of the actuator, essential to provide meaningful force feedback, 2) compact form factor, and 3) fast transformation speed for real-time interactions.
For example, the vertical load-bearing capability of ShapeBots is only 0.3N with extended states. Also,  the linear actuator can easily buckle with hand pressure~\cite{suzuki2019shapebots}.

\begin{figure}[h!]
\centering
\resizebox{\columnwidth}{!}{
\includegraphics[height=1\textwidth]{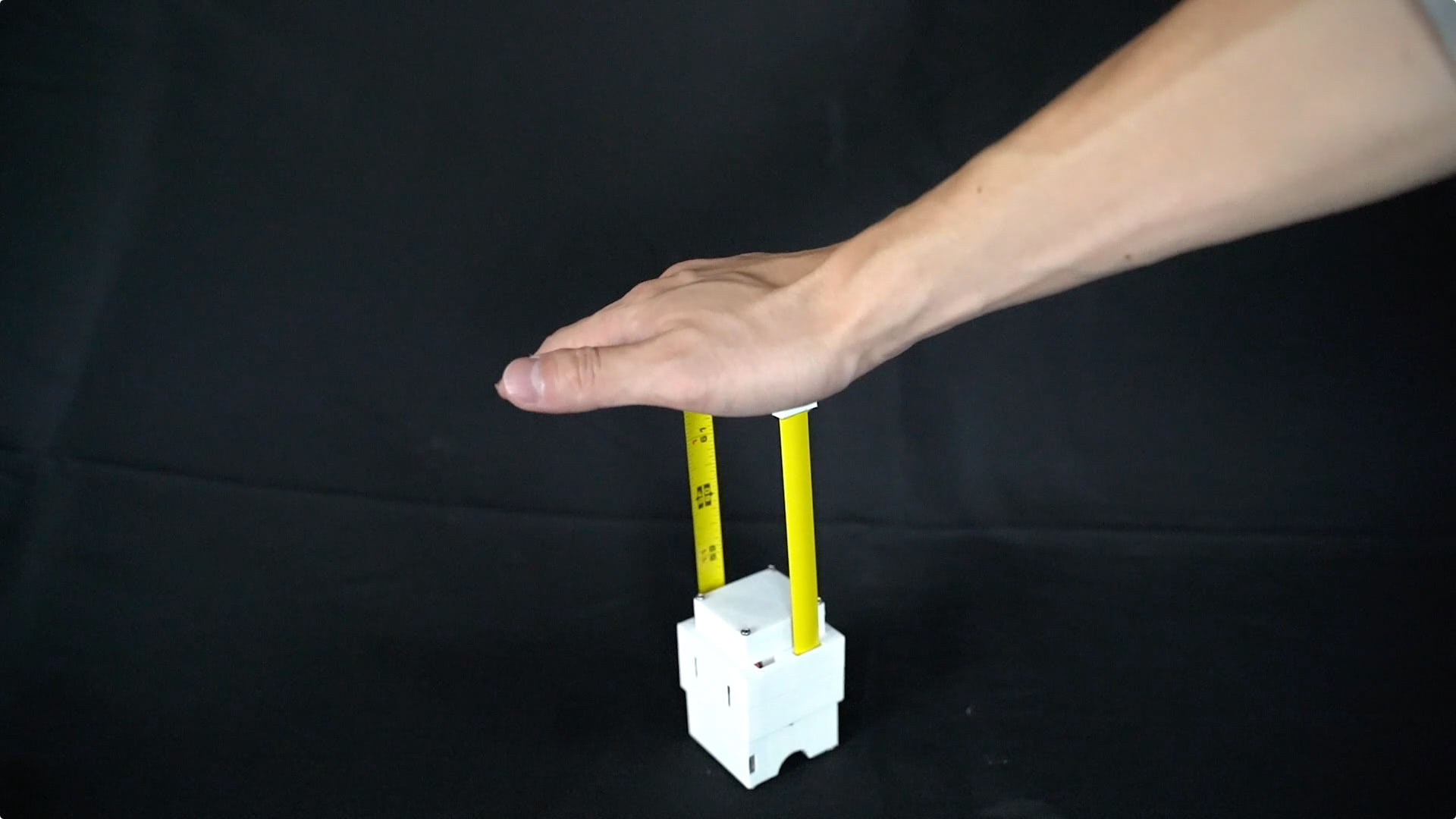}
\includegraphics[height=1\textwidth]{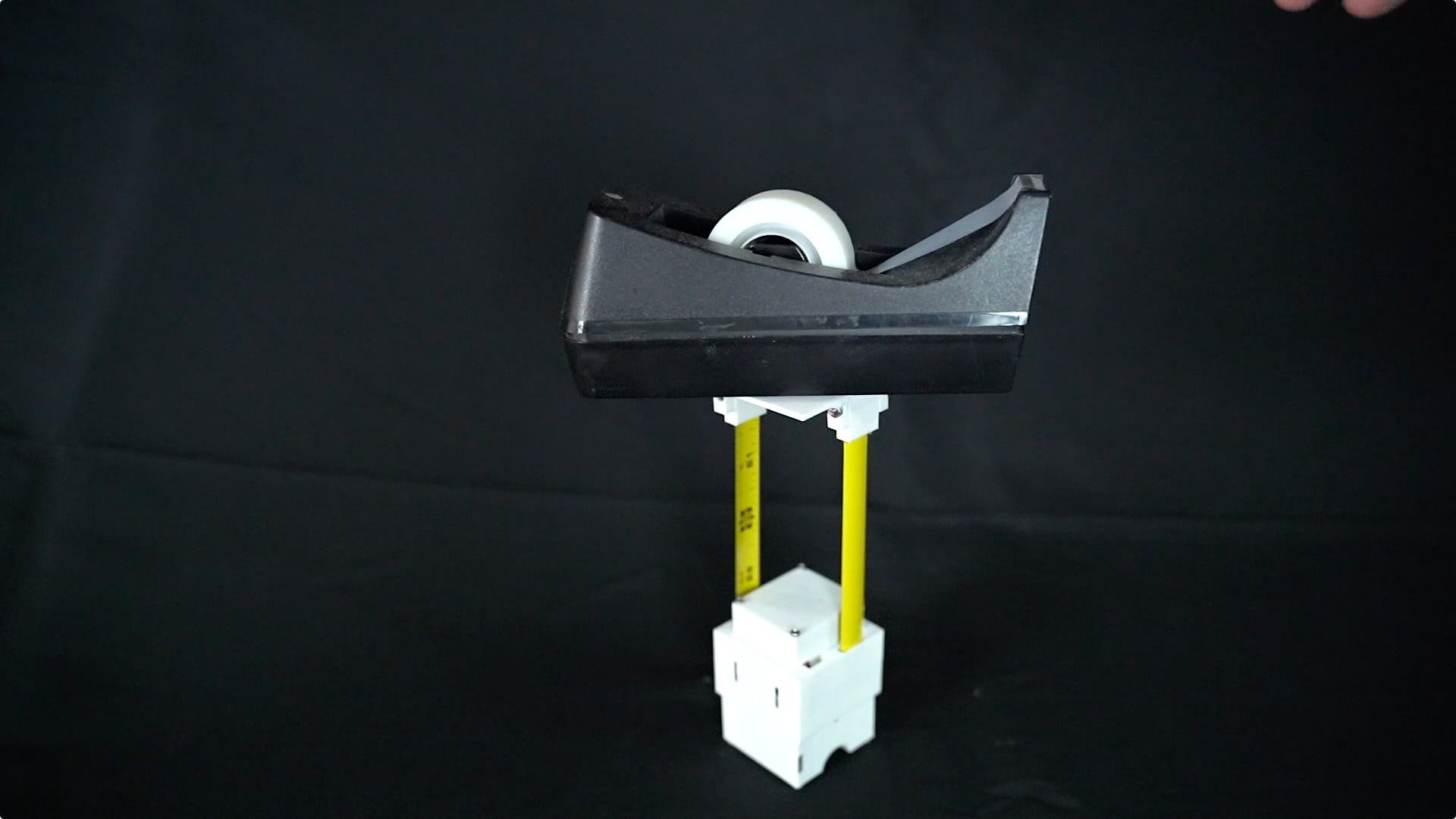}
}
\caption{The actuator maximum holding load is 1379 g at the extended state (25 cm height)}
\Description{
The actuator maximum holding load is 1379 g, demonstrated by a user resting their hand on a robot with extended actuator and by a tape dispenser on top of a robot extended actuator.
}
~\label{fig:mechanical-force}
\end{figure}

Our \system{} linear actuator is designed to achieve all of these requirements with reasonable force capabilities.
Figure~\ref{fig:mechanical-design} illustrates the mechanical design of a linear actuator. 
In our design, the two retractable metal tapes on motorized reels occupy a small footprint but extend and hold their shape while resisting modest loads in certain directions. 
Our reel-based linear actuator uses compact DC motors (Pololu 1000:1 Micro Metal Gearmotor HP 6V, Product No. 2373).
This motor has a cross-section of 1.0 $\times$ 1.2 cm and its length is 2.9 cm. 
The no-load speed of the geared motor is 31 rpm, which extends the metal tape at 2.8 cm/sec.
The motor's maximum stall torque is 12 kg$\cdot$cm.
We accommodate two motors placed side by side to minimize the overall footprint size.

\begin{figure}[h!]
\centering
\resizebox{\columnwidth}{!}{
\includegraphics[height=1\textwidth]{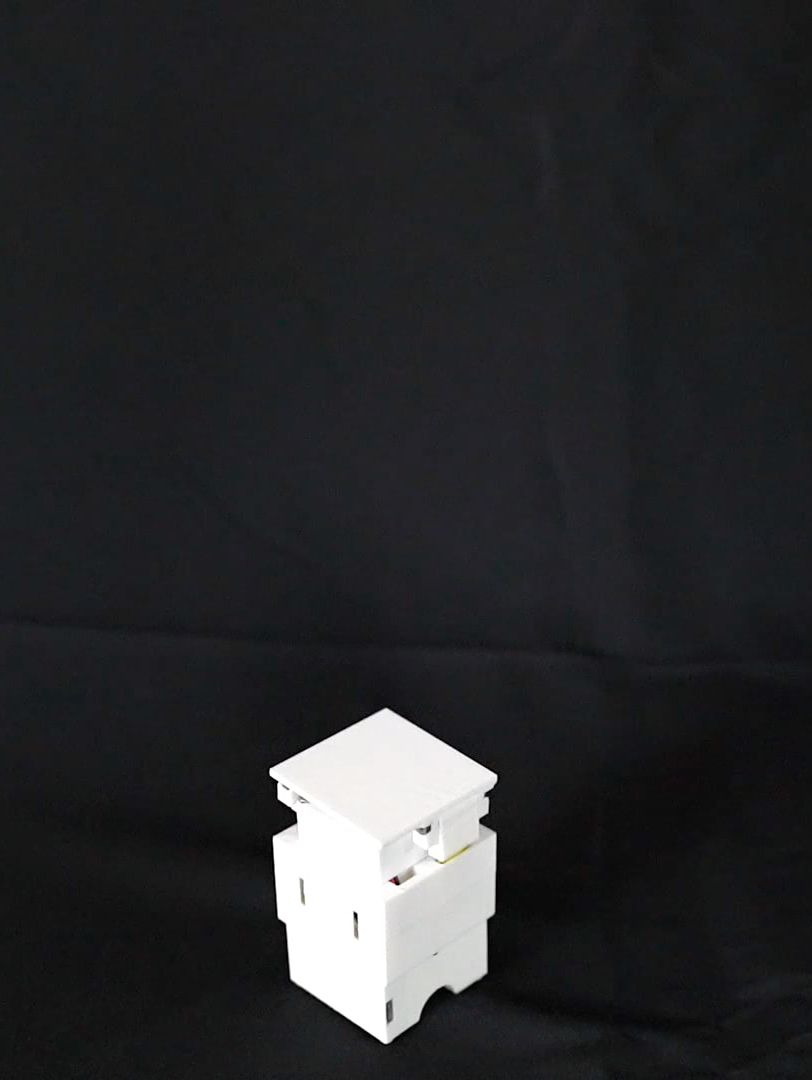}
\includegraphics[height=1\textwidth]{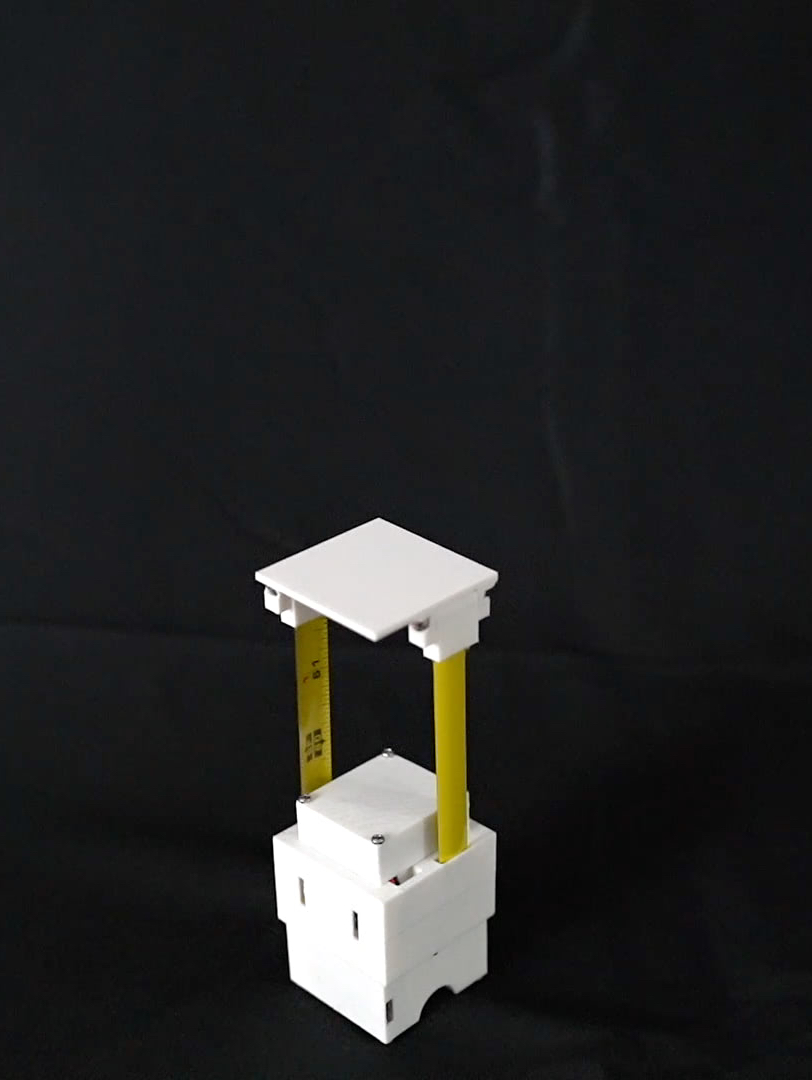}
\includegraphics[height=1\textwidth]{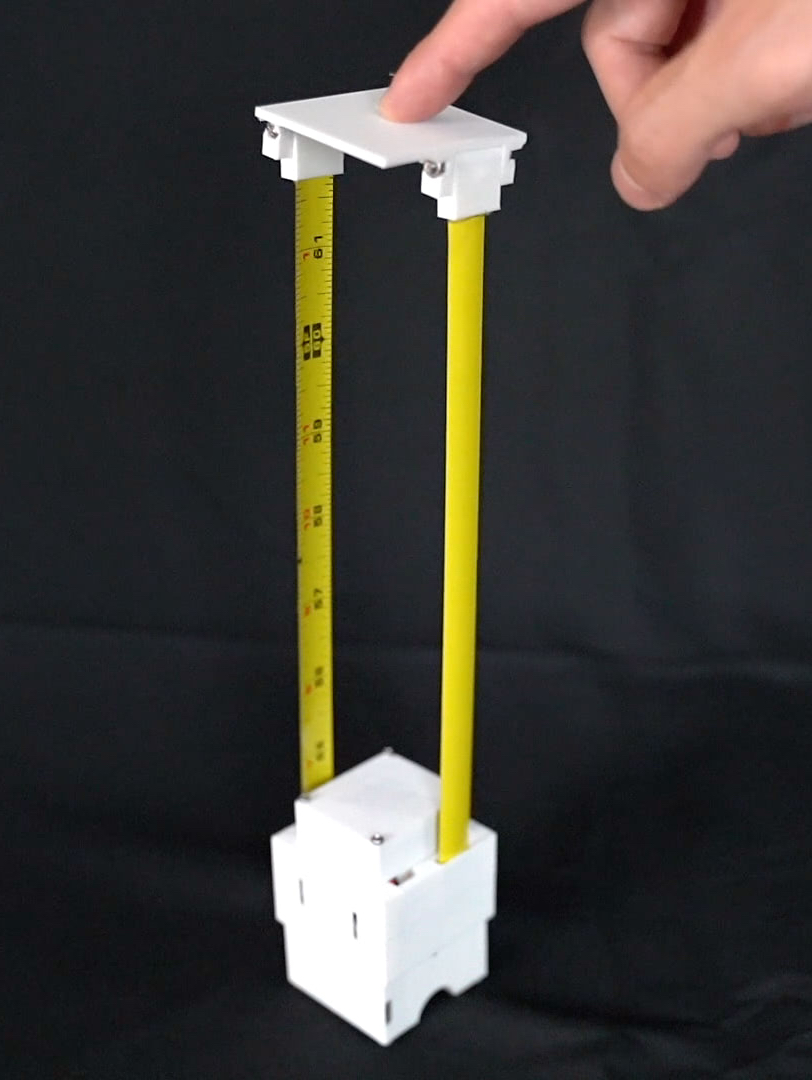}
}
\caption{The reel actuator extends the tape at 2.8 cm/sec from a minimum height of 8cm to a maximum height of 32 cm.}
\Description{
A sequence of three figures depicts that the reel actuator extends 2.8 cm/sec from a minimum height of 8cm to a maximum height of 32 cm.
}
~\label{fig:system-extend}
\end{figure}

For the reel, we use an off-the-shelf metal tape measure reel (Crescent Lufkin CS8506 1/2 x 6 inch metal tape measure).
The material choice of this reel is one of the key design considerations as it determines the vertical load-bearing capability.
On the other hand, a strong material makes it more difficult for this small DC motor to successfully rotate and rotate the reel.
After the test of eight different tape measures devices with various materials, stiffnesses, thicknesses, and widths, we determined the Crescent Lufkin CS8506 tape measure to work most reliably in our setting. 
The tape has 0.15 mm thickness and is 1.2 cm (1/2 inch) width wide, and slightly curved to avoid buckling. 
We cut this tape measure  to 36 cm and drilled a 3 mm hole at the end to fix it to the shaft with an M3 screw.

Two DC motors + gears individually rotate to extend and retract the reels. 
Each reel is connected to a tiltable planar cap.
The cover cap is made of 3D printed parts (4.7 $\times$ 4.7 cm) and has a shaft on each side fastened with an M3 screw (2.6 cm in length) and nut to make each end rotatable.
By individually controlling the extension length of each tape, the top surface can tilt between -60 to 60 degrees.
We use a rotary encoder (Pololu Magnetic Encoder Pair Kit, 12 CPR, 2.7-18V, Product No. 4761) connected to the motor shaft to continuously measure the  position of each reel and hence extension of the tape and tilt of the cap.

The overall footprint of our actuator has a cross-section of 4.7 $\times$ 4.7 cm and 3.0 cm in height.
The robot's height can change from 8 cm in minimum to 32 cm at maximum.
The no-load extension/retraction speed is 2.8 cm/sec.
The vertical load-bearing capability is approximately 13.53 N (at the extended state), which is strong enough to withstand a modest human touch force (describe more detail in the technical evaluation section).

\begin{figure}[h!]
\centering
\resizebox{\columnwidth}{!}{
\includegraphics[height=1\textwidth]{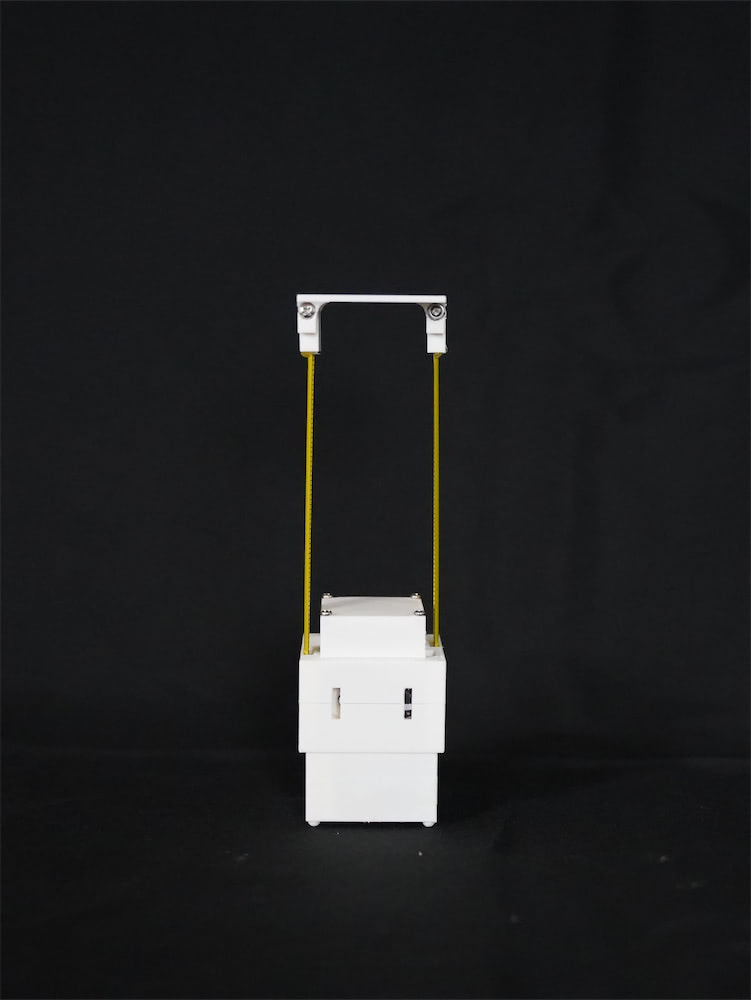}
\includegraphics[height=1\textwidth]{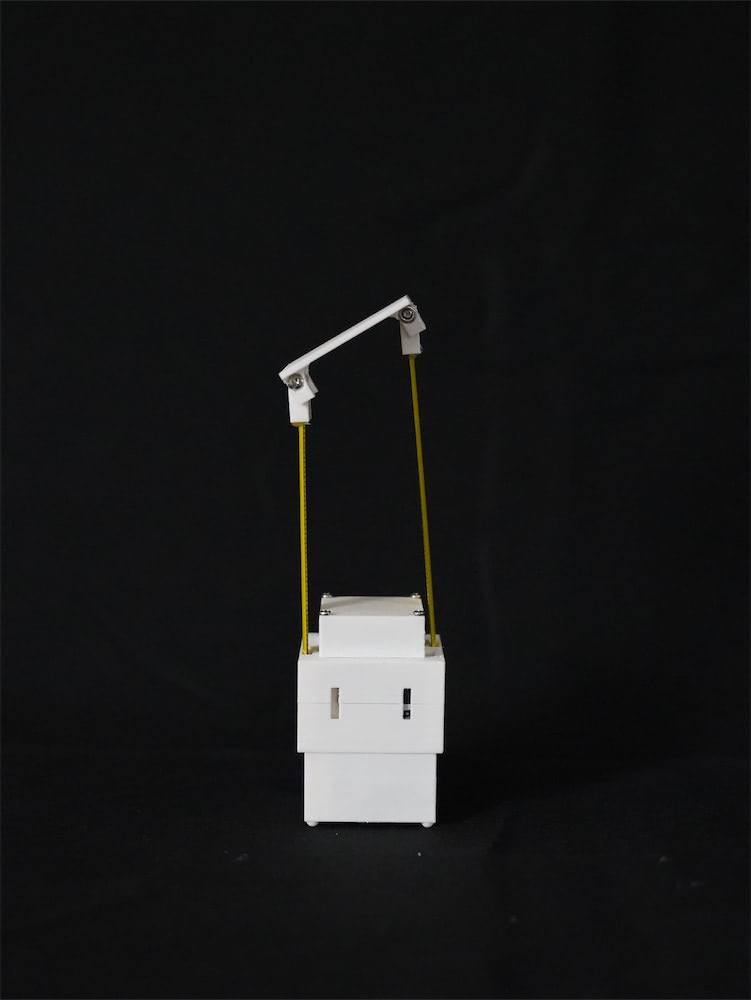}
\includegraphics[height=1\textwidth]{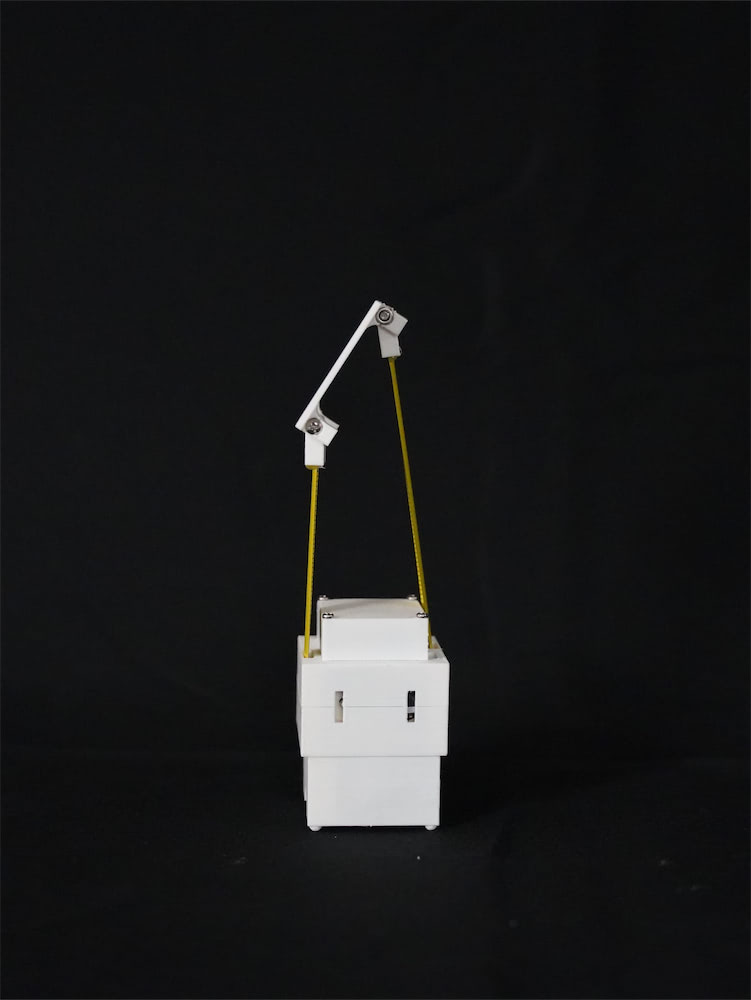}
}
\caption{The cap can tilt up to 60 degrees.}
\Description{
A sequence of three figures depicts that the reel actuator cap can tilt up to 60 degrees.
}
~\label{fig:system-extend-2}
\end{figure}

\subsubsection{Design Rationale}
During our initial prototyping phase, we have considered and prototyped various different mechanisms, including lead screws (similar to shapeShift~\cite{siu2018shapeshift}), telescopic pole (similar to motorized telescopic car antenna), and inflatable structure (similar to TilePoP~\cite{teng2019tilepop}, LiftTiles~\cite{suzuki2020lifttiles}, PuPoP~\cite{teng2018pupop}, and Pneumatic Reel Actuator~\cite{hammond2017pneumatic}).
These designs are suitable for haptic devices as they can withstand the strong vertical force (e.g., pushing force with a hand), but they often require a longer vertical travel and lower extension ratio (e.g., lead screw, telescopic pole) or slow transformation speed and complex control system (e.g., pneumatic actuation).
Therefore, we decided to explore and develop the current design to meet our design requirements.

\subsubsection{Electronic Design}
\begin{figure}[h!]
\centering
\resizebox{1\columnwidth}{!}{
\includegraphics[width=0.48\columnwidth]{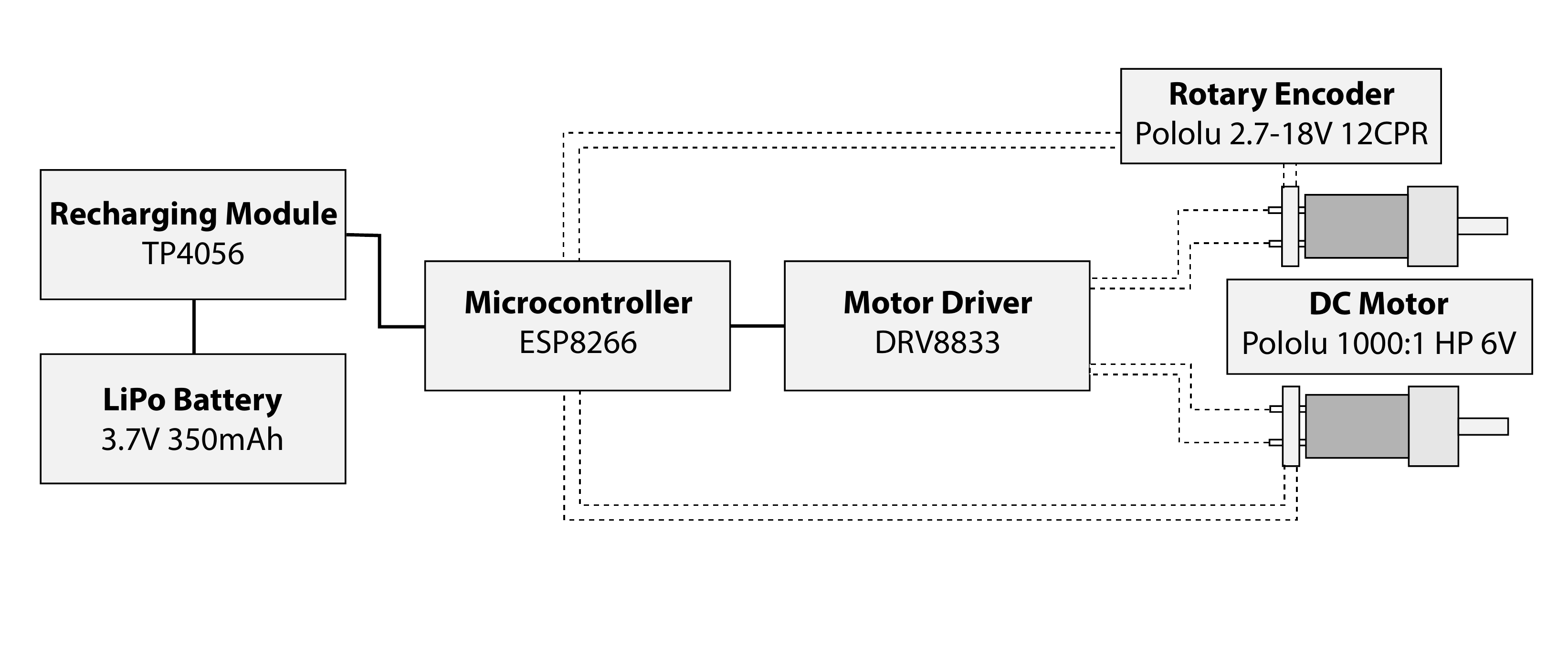}
}
\caption{System schematics of the actuator electronics.}
\Description{
System schematic of the actuator electronics.
}
~\label{fig:electronics}
\end{figure}

Figure~\ref{fig:electronics} illustrates the schematic of \system{}' electronic components. 
We use an ESP8266 microcontroller (Wemos D1 mini, 3.4 $\times$ 2.5 cm) to control two motors, read signals from two rotary encoders, and communicate with the main computer through Wi-Fi communication with a user datagram protocol (UDP).
Each module connects to a private network and is assigned a unique IP address. 
The computer sends a target height to each IP address, and the microcontroller controls the rotation of the motor by measuring the rotation count based on the rotary encoder.
The microcontroller controls one dual motor driver (Pololu DRV8833 Dual Motor Driver Product No. 2130), which can control two DC motors independently.
The operating voltage of all modules is 3.5V and the microcontroller is connected to 3.7V LiPo battery (350mAh 652030) through a recharging module (TP4056). 
The module is rechargeable through an external micro-USB cable.

The entire linear actuator component measures 4.7 $\times$ 4.7 cm $\times$ 8.0 cm.
Our prototype costs are approximately \$70 USD for each actuator, including the microcontroller (\$3 USD), motors (\$20 USD $\times$ 2), a motor driver (\$4 USD), rotary encoders (\$7 USD $\times$ 2), a battery (\$6 USD), a custom PCB (\$1 USD), and a tape measure (\$6 USD), but the cost could be significantly decreased with mass production and using alternative low-cost motors.

\subsection{Mobile Robot Base}
\begin{figure}[h!]
\centering
\resizebox{1\columnwidth}{!}{
\includegraphics[width=\columnwidth]{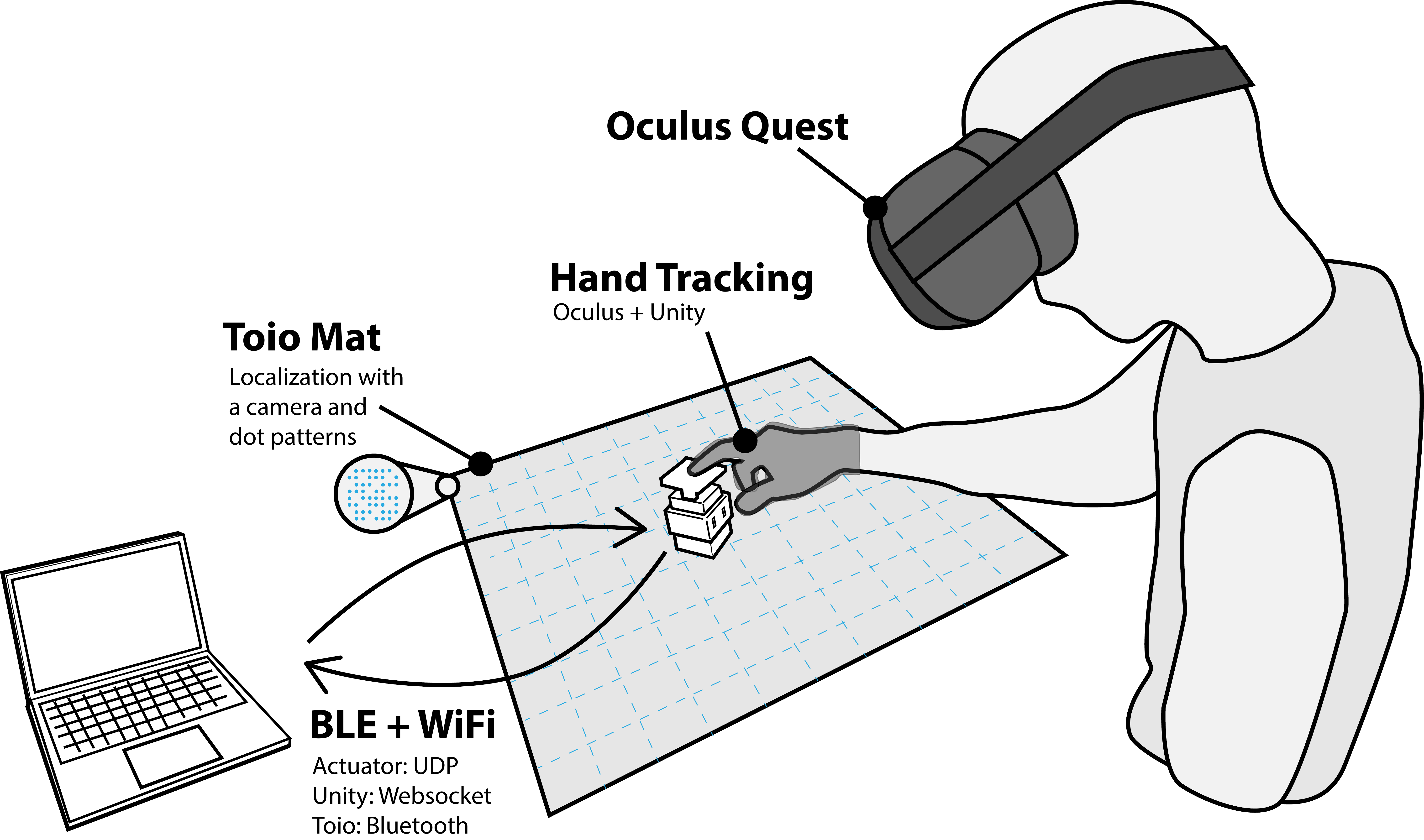}
}
\caption{System Setup: The Oculus Quest VR Headset is calibrated to work in the same coordinate space as the tracked robots.}
\Description{
System setup consisting of a laptop, a Toio mat, a HapticBot and a user wearing an Oculus Quest VR headset and touching the HapticBot. The laptop communicates with the robot via BLE and WiFi.
}
~\label{systemsetup}
\end{figure}

As a mobile robotic base, we use an off-the-shelf mobile robot (Sony Toio\textsuperscript{TM}) that features two-wheeled robots and Bluetooth control.
The main reason we chose the Toio robot is its sophisticated and easily deployable tracking system. In addition, Toio robots have numerous advantages in 1) off-the shelf availability, 2) light and compact,  3) fast, 4) fairly strong, and 5) safe. 
For tracking and localization, Toio has a built-in look-down camera at the base of the robot to track the position and orientation on a mat by identifying  unique printed dot patterns, similar to the Anoto marker~\cite{faahraeus2007electronic}. The built-in camera reads and identifies the current position of the robot, enabling easy 2D tracking of the robots with no external hardware (Figure~\ref{systemsetup}).

We use a publicly available JavaScript API to programmatically track and control the robots from a computer ~\footnote{\url{https://github.com/toio/toio.js}}.
The size of each robot has a cross-section of 3.2 $\times$ 3.2 cm and 2.5 cm in its height. 
The tracking mat (Toio Tracking Mat\textsuperscript{TM}) has 55 $\times$ 55 cm of covered area, but the interaction area can be easily extended with multiple mats.
The resolution of 2D position and orientation detection are 1.42 mm and 1 degree, respectively.
The Toio robot alone (without the shape changing module) can drive and rotate with the maximum speed of 35 cm/sec and 1500 degrees/sec, respectively.
The maximum load-bearing capacity of Toio robot is 200g, but can move heavier objects with an appropriate caster base that can evenly distribute the weight. Each robot costs approximately \$40 USD.

\subsubsection{Path Planning and Control Mechanism}
\begin{figure}[h!]
\centering
\resizebox{\columnwidth}{!}{
\includegraphics[height=1\textwidth]{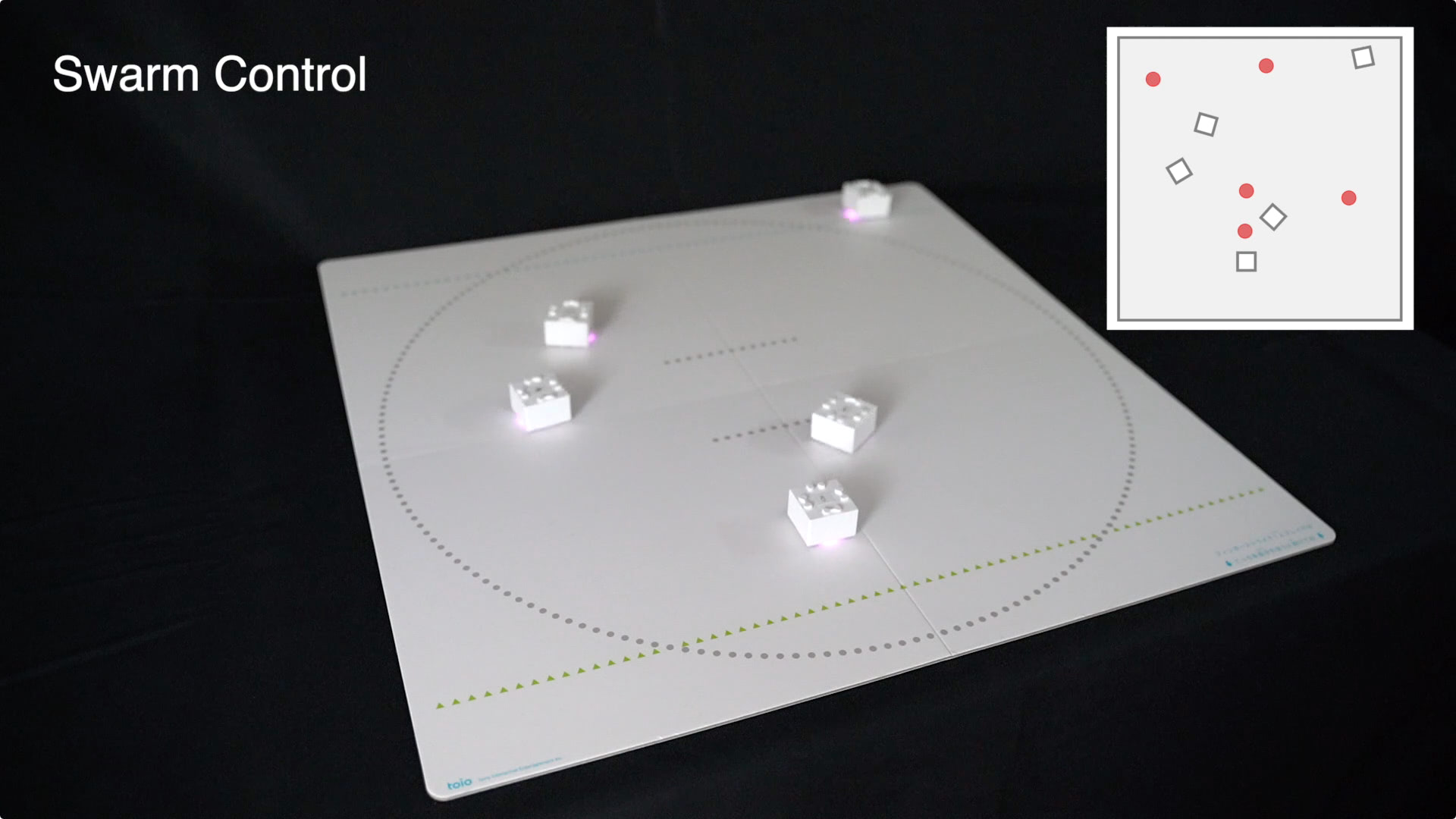}
\hspace{0.3cm}
\includegraphics[height=1\textwidth]{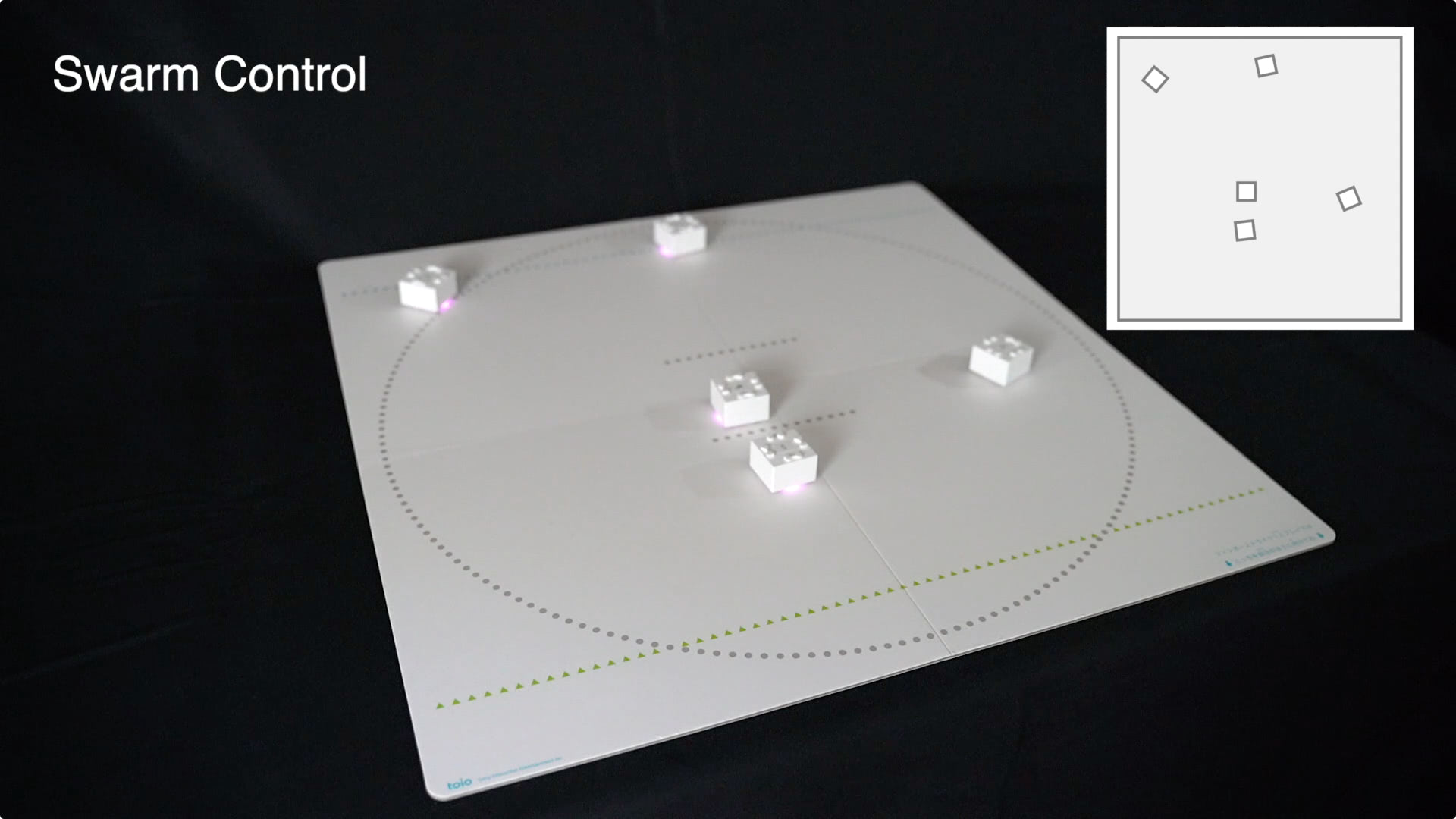}
}
\caption{Our path planning and control algorithm coordinates multiple robots.}
\Description{
Our path planning and control algorithm coordinates the robot swarm, depicted in two side by side figures with 5 robots on a tracking mat, and little inset figures that render a view of the software.
}
~\label{pathplanning}
\end{figure}

Based on Toio's internal tracking mechanism, the software reads and controls the position and orientation of the multiple robots simultaneously (see Figure~\ref{pathplanning}). 
To achieve this, we adapt the Munkres algorithm for target assignment and Reciprocal Velocity Obstacles
(RVO) algorithm~\cite{van2008reciprocal} for collision avoidance. 

Given the current position of each robot and its target position, the algorithm first assigns each target by minimizing the total travel distance of all robots.
Once a target is assigned, the system navigates the robot until all of the robots reach to their target positions, while avoiding the collisions with RVO.
The driving speed dynamically changes based on the distance between the current and target position, to maximize speed when the target is far and slows it down when approaching to the target to reduce the braking distance and avoid overshooting.
The distance to stop a moving robot is 2 mm and 5 degrees in orientation.

\subsection{Software Implementation}

\subsubsection{Software Architecture}
For the virtual reality environment and gestural tracking, we use an Oculus Quest HMD with its built-in hand tracking mechanism (see Figure~\ref{systemsetup}).
The main computer (MacBook Pro 16 inch, Core i9, 32GB RAM) runs as the Node.js server.
The server controls and communicates with the linear actuators and the Unity application on an Oculus Quest through Wi-Fi (with UDP and Websocket protocol, respectively) and Toio robots through Bluetooth.
The host computer communicates with seven Toio robots through Bluetooth V4.2 (Bluetooth Low Energy).
The \system{} computer operates at 60 FPS for tracking and control.

\subsubsection{Approximating a Virtual Surface in Unity}
We use the Unity game engine to render a virtual environment. As each robot moves along the planar surface, it constantly changes its height and orientation to best fit the virtual surface above it.
To obtain the target height and surface normal of the robot, the system uses a vertical ray casting to measure the height of the virtual contact point given its position on the desk.

Given the virtual object or surface, and the robot position in Unity space  $(x, y, z_0)$, where $z_0$ is the height of the table, we cast a ray vertically from a high elevation (above the height of the virtual geometry) $(x, y, z_0 + H)$ where $H$ is 1 meter, down until it intersects the virtual surface at $(x, y, z)$. We obtain the distance until the ray hits to the surface $0 \leq d_{(x, y)} \leq H$, thus we get the height at which the HapticBot needs to render the virtual surface, with $height_{(x, y)} = z - z_0 = H - d_{(x, y)}$. 
The ray casting is preformed from above and not from the robot height, to avoid culling effects, as the virtual geometry is facing up toward the user.
To obtain the tilting angle of the robot's top surface (e.g., to render a tilted roof of a house), we cast two rays, each at the locations of each actuator attachment to the plane, and change the actuator heights accordingly.

\subsubsection{Target Assignment for Haptic Interactions}
We perform this height measuring at each 0.5 cm grid point of the mat (55 cm $\times$ 55 cm) at every 0.5 seconds.
Based on this, the system obtains the height map of the playground area.
With this, the system can identify regions that the robot should provide a touchable plane (i.e., the region that has a surface higher than a certain height, in our setting where 1 cm is the threshold).
The system sets these regions as target objects or surfaces, then moves the robot within this target region while the user moves her finger laterally. 
When the number of regions exceeds the number of robots, we optimize the target position based on the finger position.
For example, when the robots need to cover four separate target regions, the robots move across the region that is closest to the current finger position. 
Multiple robots can also support a single large continuous region when there is enoughof robots.
This way, we can assign the target position to each of virtual objects (e.g., virtual houses, buildings, and cars on a map), either they are static (e.g., buildings) or dynamic (e.g., moving cars).
By leveraging the combination of hand tracking and dynamic target assignment, the smaller number of robots can successfully simulate the virtual haptic proxy.

\begin{figure}[h!]
\centering
\includegraphics[width=1\columnwidth]{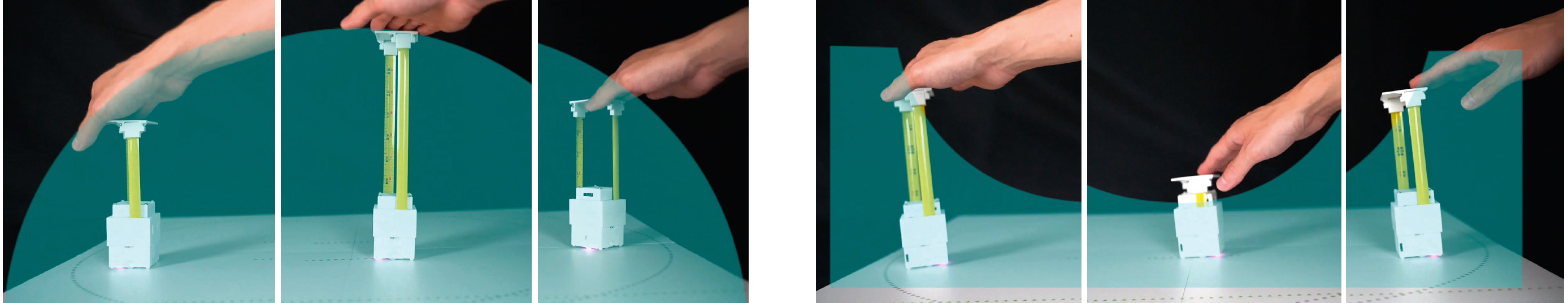}
\caption{A single robot simulates a virtual surface.}
\Description{
Left: Side view of a sequence of three shots of a single robot moving to simulate a virtual sphere while the user touches the top cap.
Right: Side view of a sequence of three shots of a single robot moving to simulate a virtual concave shape while the user touches the top cap.
}
~\label{fig:simulatingsphere}
\end{figure}

\subsubsection{Finger Tracking and Continuous Positioning}
To enable the robot to encounter the user's finger whenever she wishes to touch a virtual surface, a camera  tracks the position of the user's finger, and moves the robot to minimize the robot's distance to the finger.
We use the built-in hand tracking of the Oculus Quest that generates the user's finger position inside the Unity virtual space.
To do that we need to be able to generate a reference between Unity coordinate system, and Toio's mat reference. We use a simple calibration process to match the coordinate systems between Toio mat and the virtual floor in the Unity. 
\diff{The user calibrates the mat through the following two steps. First, the user places her index finger at the center of the mat and presses a space key then the center of the virtual mat becomes the location of the index finger. Second, the user places her index finger on the left bottom corner and presses the space key again, then the virtual mat rotates to match with the physical mat. Although this brief calibration routine is needed to synchronize the Toio and Oculus coordinate systems, it does not limit the deployability and portability---the user can pack all parts of the system in a backpack, go to a new room, and set it up there with minimal effort. 
}


The Oculus tracks the user's hands at 60Hz and the target position is transmitted to the corresponding robot. The system has enough bandwidth to support multiple robots (we use seven robots in total).
By leveraging the combination of hand tracking and dynamic target assignment, a small number of robots can simulate a large haptic surface.

\subsection{Technical Evaluation}

We evaluated the technical capabilities of \system{} in terms of the following aspects. The table summarizes the results of measurements for each of them.

\setlength{\tabcolsep}{5pt}
\renewcommand{\arraystretch}{1.1}
\begin{tabular}{ l|l }
\small 1-a) Maximum Vertical Load Holding (at 12 cm height) & 21.8 N \\
\small 1-b) Maximum Vertical Load Holding (at 25 cm height) & 13.5 N \\
\small 1-c) Maximum Vertical Load Holding (tilted at 25 cm height) & 9.31 N \\
\small 2) Maximum Vertical Load Lifting & 10.8 N \\
\small 3) Maximum Horizontal Pushing Force & 0.31 N \\
\small 4) Speed of Vertical Extension & 2.8 cm/s \\
\small 5) Speed of Horizontal Movement & 24 cm/s \\
\small 6-a) Average Target Reach Time (One Robot)    & 2.0 sec \\ 
\small 6-b) Average Target Reach Time (Three Robots) & 1.7 sec \\ 
\small 6-c) Average Target Reach Time (Seven Robots) & 1.6 sec \\ 
\small 7) Average Vertical Extension Error & 3 mm \\
\small 8) Average Tilting Angle Error & 5 deg \\
\small 9-a) Average Position Error & 3 mm \\
\small 9-b) Average Orientation Error & 3 deg \\
10) Latency & 80 ms
\end{tabular}

\subsubsection{Method}
1-3) We employed a Powlaken's food scale (accurate to 0.1g) for force measurements and analyzed the number via video recordings. For the load-bearing, we put the robot on top of the scale and pushed down with hands until the reel buckles. For the maximum force of vertical extension, we place a fixed surface above the robot so that it can push against it while sitting on on the scale that measures the vertical force plus weight of the robot (which is subtracted). 
Then, the robot extends the actuator and measure the force to push the object through the reaction force measured with the scale on the bottom. For the horizontal pushing force, we place the scale horizontally and let the robot bump and push the scale. We measured the maximum impact force against the scale.
4-5) We measured the speed of the robot and linear actuator via video recordings of the robot moving next to a ruler.
6) We measured the time to reach the target via video recordings. The target points are randomly assigned and measured the time until the last robot reaches the target. We evaluated this with one, three, and seven robots.
7-8) For the extension and tilting accuracy of the linear actuator, we took three measurements of the extension length given the same motor activation duration for three different duration values.
9) For position and orientation accuracy, we logged the distance and angle deviation between the robot and its target with our control system, given a certain duration.
10) We measured the latency of each step (i.e., Wi-fi  communication, tracking, and computation of path planning) by comparing timestamps from the start of each event to its conclusion with four attempts for five robots. 


\subsubsection{Results}
The robot's vertical load-bearing capability is 13.53 N (at 25 cm height) and load lifting capability is 10.08 N.
This is a significant improvement of the existing similar work (e.g., 0.3 N in ~\cite{suzuki2019shapebots}), which is important for haptic devices.
The maximum speed of the vertical extension was 2.8 cm/sec.
The average target reach time ranges from 1.6 to 2.0 sec with the maximum speed of the robot's horizontal movement was 24 cm/sec.
This is slightly slower than an unmodified Toio robot due to the added weight.
We found that the moving robot can deliver a maximum force of 0.31 N for horizontal pushing force.
The measured average positional error for vertical extension and horizontal movement were both 3 mm.
Finally, the measured latency was 80 ms for the total latency of the control loop with three robots.

\section{User Evaluation}
We conducted a user study to evaluate the \system{} ability to simulate: (1) haptic encounters of different types, (2) continuous touch on different tilts. In our evaluation we aim to compare the performance of \system{} rendering continuous surfaces to a shape display.
To do so we designed 2 tasks and recruited 6 participants for each task. Totalling 12 participants ages 18 to 50 (8 female). \diff{Most of our participants did not have any prior experience with haptic devices.}  After their corresponding task, participants completed an offline questionnaire.

\subsection{Task 1: Haptic Encounters}

In this task, we evaluate the ability to render various objects of different size, form, and texture by a shape changing robot.
\diff{We are interested in how HapticBots can ``approximate'' some of the presented objects.}
To do so, we introduce a larger range of objects with different convexity and/or texture, including a mag cup, a rubik's cube, a tennis ball, a wrench, and a sea urchin.

\begin{figure}[h!]
\centering
\resizebox{\columnwidth}{!}{
\includegraphics[height=1\textwidth]{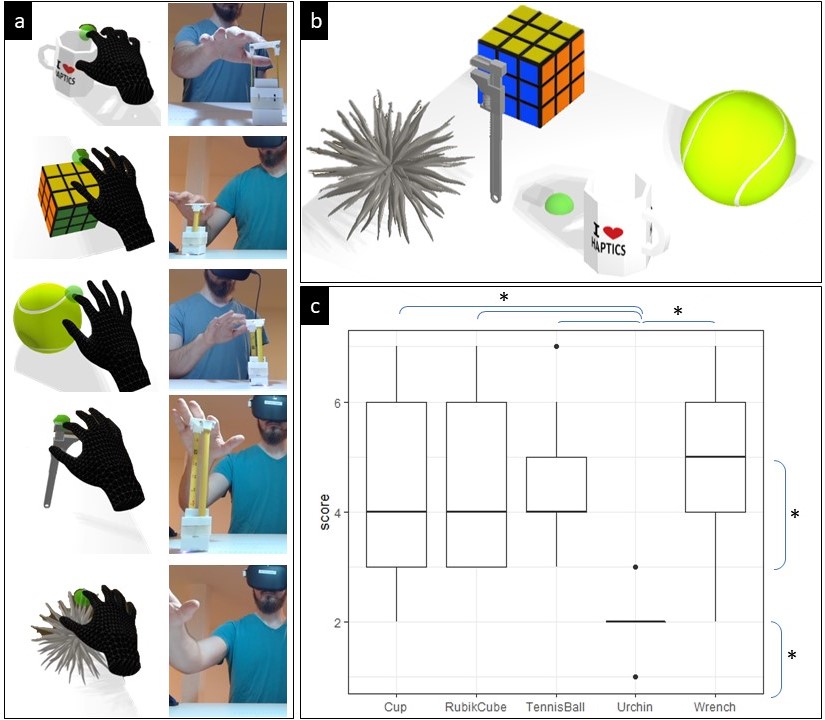}
}
\caption{a) Participant in the haptic encounter study reaching for the 5 different objects. b) The five objects that they encounter in different areas of the study table and sizes. c) Boxplot of the realism score results. Marked with an asterisk significant differences. The thick lines are the medians, and the boxes are the interquartile ranges (IQR). The whiskers follow the standard convention of extending to 1.5 times the IQR or the maximal/minimal data
point. }
\Description{
a) Participant in the haptic encounter study reach-ing for the 5 different objects. b) The five objects that they encounter in different areas of the study table and sizes. c)Boxplot of the realism score results. Marked with an asterisk significant differences. The thick lines are the medians, andthe boxes are the interquartile ranges (IQR). The whiskers follow the standard convention of extending to 1.5 times theIQR or the maximal/minimal data point.
}
~\label{study1objects}
\end{figure}

We ask participants to rate their experience in response to the question: "How realistic did the touch encounter feel?" from 1 to 7.
Participants in this task were requested to only touch the top of the appearing object for 10 seconds with one finger, but once touched, they were free to move their finger over the surface and the robot which moved to be under the finger tip and adapt the height to the virtual object (see supplementary video).
Each of the five objects appeared two times and the average score was computed for each object (Figure \ref{study1objects}).

\subsubsection{Results}
Overall, the results (Figure \ref{study1objects}) show that all objects could be realistically rendered by \system{} during the encounters. And all of them, minus the spiky urchin, scored significantly higher than 3 (Wilcoxon signed rank test with continuity correction V=26, p<0.05), the mean score was $4.55 \pm 1.4$ SD. The spiky urchin that was significantly lower than all the rest (V=36, p<0.01), which scored $2 \pm 0.7$ SD, and despite the very non-compatible shape of the object scored significantly higher than 1 (V=28, p=0.01).

\subsection{Task 2: 3D Continuous Touch}

In this task, we evaluate the ability to render a large continuous surface of the different tilt angles.
To evaluate this, we presented participants with a total of five different tilted virtual surfaces of 0, 15, 30, 50 and 70 degrees of inclination. 
Then, we measured how precisely the participants can recognize each tilt angle.

The second task's motivation is to evaluate the capability of large surface rendering with a single robot (e.g., Section 3.2.1), particularly compared to the other state-of-the-art encountered-type approach.
To date, the only approach that can dynamically render a large haptic surface is shape displays (see Related Work); thus we aimed to compare the performance with the shape display approach.
However, most large pin-based shape displays are rare due to their high cost, making it infeasible to replicate or prepare the fully functional shape display only for this task.
Thus, we built a wooden mock-up shape display for our base condition to simulate the shape display's haptic sensation.
Our mock-up have a stair shape instead of a slope, is to mimic the discrete pins of the shape display. The resolution of our mock up shape display is 1.5 cm, following prior works~\cite{follmer2013inform, iwata2001project, nakagaki2019inforce}.
We only used the \system{}' lateral motion For fair comparison and did not use the tilt functionality.

\begin{figure}[h!]
\centering
\resizebox{1\columnwidth}{!}{
\includegraphics[height=1\textwidth]{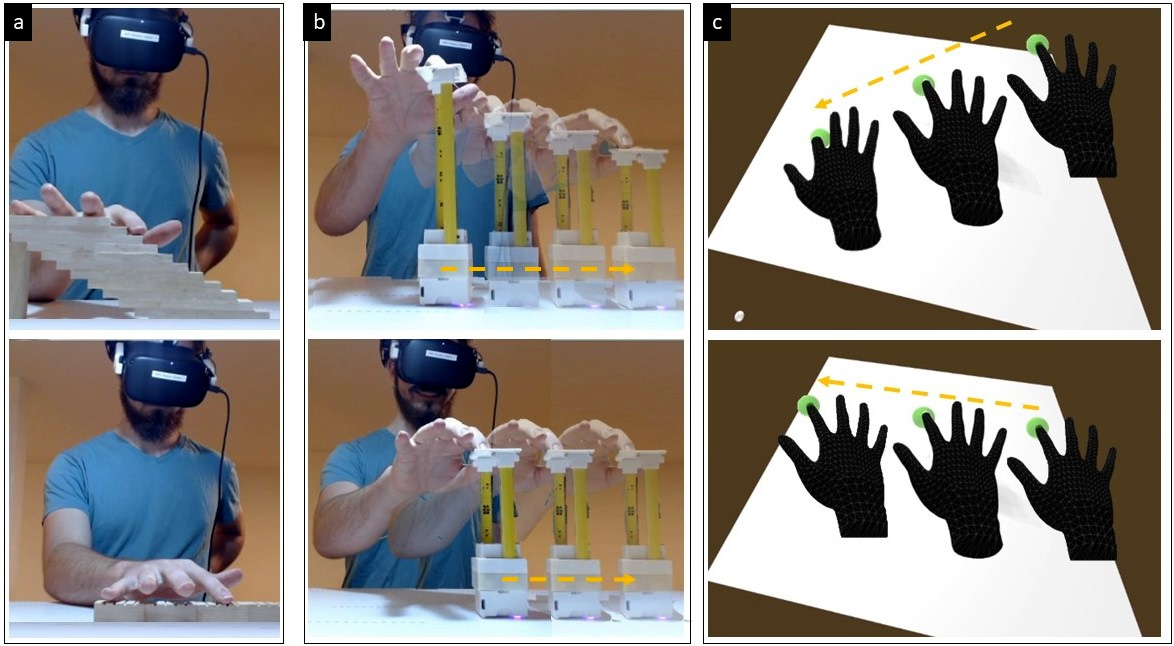}
}
\caption{a) Participant assessing the tilt on a static shape display made of wooden bricks. b) Participant assessing the tilt using \system{}. c) First Person Perspective inside the HMD during the tilt estimation task. 
}
\Description{
a) Participant assessing the tilt on a static shape display made of wooden bricks. b) Participant assessing the tilt using HapticBots. c) First Person Perspective inside the HMD during the tilt estimation task.
}
~\label{study2tilt}
\end{figure}

The users, while in VR, could not see the tilt but only a green dot representing the position they needed to touch with their fingers during the task. This green dot appeared in their HMD that they needed to keep following and touching in order to advance in the user test.
At the end of each tilted surface, participants had to guess the inclination and then answer the question "How much did it feel like you were touching a continuous line?" from 1 to 7. 

\begin{figure}[h!]
\centering
\resizebox{\columnwidth}{!}{
\includegraphics[height=1\textwidth]{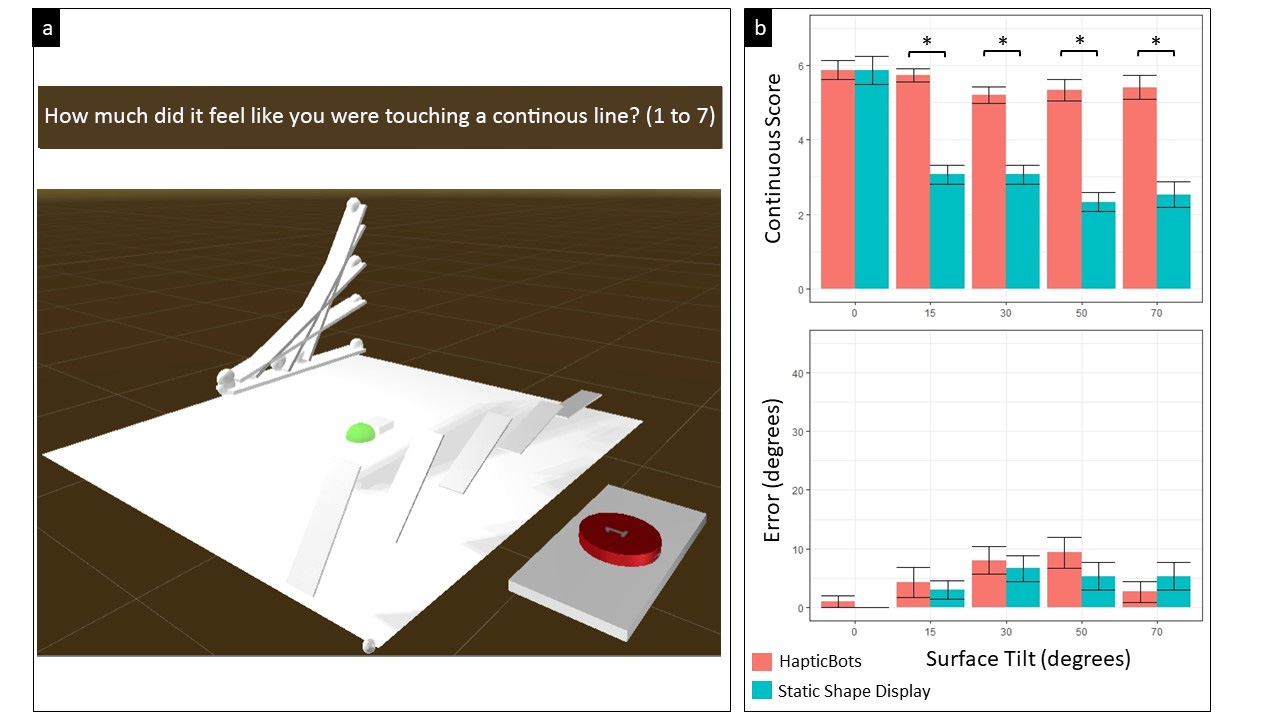}
}
\caption{a) VR evaluation. After performing the continuous touch participants had to select the tilt from a menu of tilt options and then respond to the continuity question. b) Bar charts with the scores for the perceived continuity and tilt error plots with \system{} and static shape display (error bars represent Standard Deviation).}
\Description{
a) VR evaluation. After performing the continuous touch participants had to select the tilt from a menu of tilt options and then respond to the continuity question. b) Bar charts with the scores for the perceived continuity and tilt error plots with HapticBots and static shape display (error bars represent Standard Deviation).
}
~\label{study2tiltresults}
\end{figure}

\subsubsection{Results}
We compare the results fortwo scenarios: using \system{} and using a mock-up shape display made of wood (Figure \ref{study2tilt}). 

\textbf{Tilt Estimation:} Regarding the tilt estimation, neither the repeated measures ANOVA nor the pairwise Wilcoxon signed rank test with continuity correction showed significant differences between the responses with \system{} or the static shape display (V=7, p=0.5). Post-hoc analysis per each of the different tilts also revealed no differences per any particular tilt, and none of the errors was greater than 15$^{\circ}$. Indeed, the error was quite low with both devices and participants only got $3.6 \pm 0.6$ SD errors out of the 15 tilts in the static shape display condition, and $ 3.9\pm 0.75$ SD in the \system{} condition. This shows the task was equivalently difficult to perform in both the static shape display and the \system{} and that it was well designed to not produce ceiling effects as there were still some errors in perception for about 20\% of the trials. Furthermore, the fact that the \system{} tilt perception error wasn't different than the error perceived in a real physical tilt shows the quality of control achieved on the \system{} for the task. 

\textbf{Continuity Score:} For the flat surfaces (0 degrees), both conditions performed very high on the continuity score. However, the continuity score rapidly deteriorated for the static shape display whereas \system{}  maintained a good performance across the different tilts. A repeated measures ANOVA showed that both condition and tilt were significant main effects (F(1,4)=73.9, p=0.001 and F(4,16)=12.5, p<0.0001 respectively). And that there was an interaction effect between condition and tilt (F(4,16)=6.031, p=0.004) on the score. 
In the post-hoc analysis we confirmed the interaction, there was a significant drop in continuity scores for the static shape display 
(Wilcoxon signed rank test V = 0, p = 0.001). While, no significant drop was found for the \system{} tilt estimation.

\subsubsection{Discussion}
\diff{The results show that HapticBots can be used in dynamic continuous mode without reduced slope detection compared to a solid slope of a static shape display. 
The correct classification of tilts for the HapticBots condition also shows that the presented slopes were overall distinguishable between them.
However, there are two limitations in this study.
First, since the sample size of our user study is relatively small, a further in-depth user study may be required to fully evaluate our approach.
Second, our user study only used the mock-up, static shape display and did not compare with fully-functional shape displays or robotic arms.
Since the development of these functional systems goes beyond the scope of this paper, we are interested in further evaluating our approach as future work.
}
\section{Application Scenarios}

\subsection{Education and Training}
VR is an accessible way to create realistic training setups to improve skills or prepare for complex situations before they happen in real life. With its fast encounter-type approach, users of \system{} can train their muscle memory to learn where different physical elements such as the interface of a flight cockpit are located (Figure \ref{fig:apps}).
\system{} can simulate continuous surfaces, and the robots can follow the user's fingers as they move and even elevate them during palpation diagnostics. These features could be  relevant for medical education and surgery training (Figure \ref{fig:apps}). 

\begin{figure}[h!]
\centering
\includegraphics[width=\columnwidth]{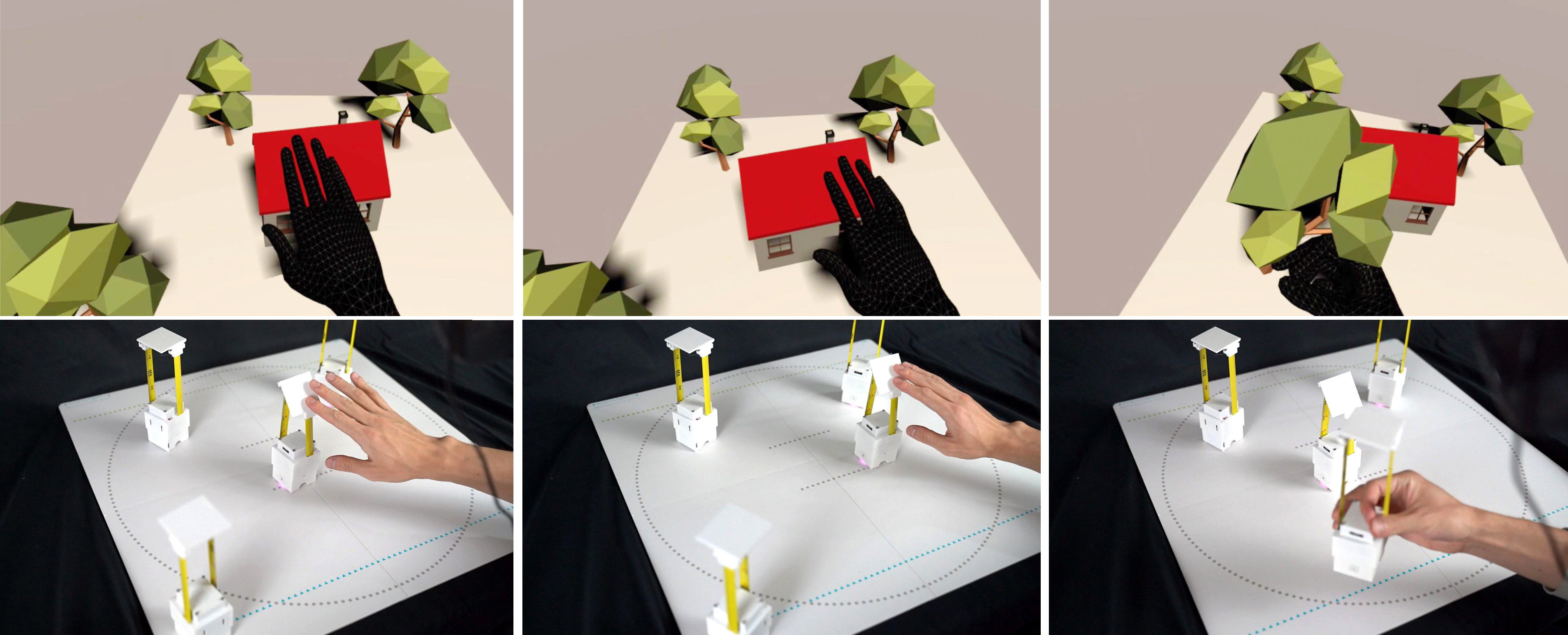}
\caption{Modeling a virtual environment: Users touch and rearrange 3D models by handling the robots.
}
\Description{
Playing chess with HapticBots, shown in a sequence of three panels with the user’s virtual view, and below it a sequence of three panels showing the HapticBots and hands
}
~\label{fig:application-house}
\end{figure}

\subsection{Design and 3D Modeling}
In addition to its continuous shape rendering capabilities, the design of \system{} being based on dual actuators makes the system robust to lateral bending and provides the ability to control different tilts to render topography of a terrain surface. This enables activities like map and city exploration or terrain simulation, which can be necessary for architectural design or virtual scene/object modeling (Figure \ref{fig:application-chess}).

\subsection{Remote Collaboration}
\begin{figure}[h!]
\centering
\includegraphics[width=\columnwidth]{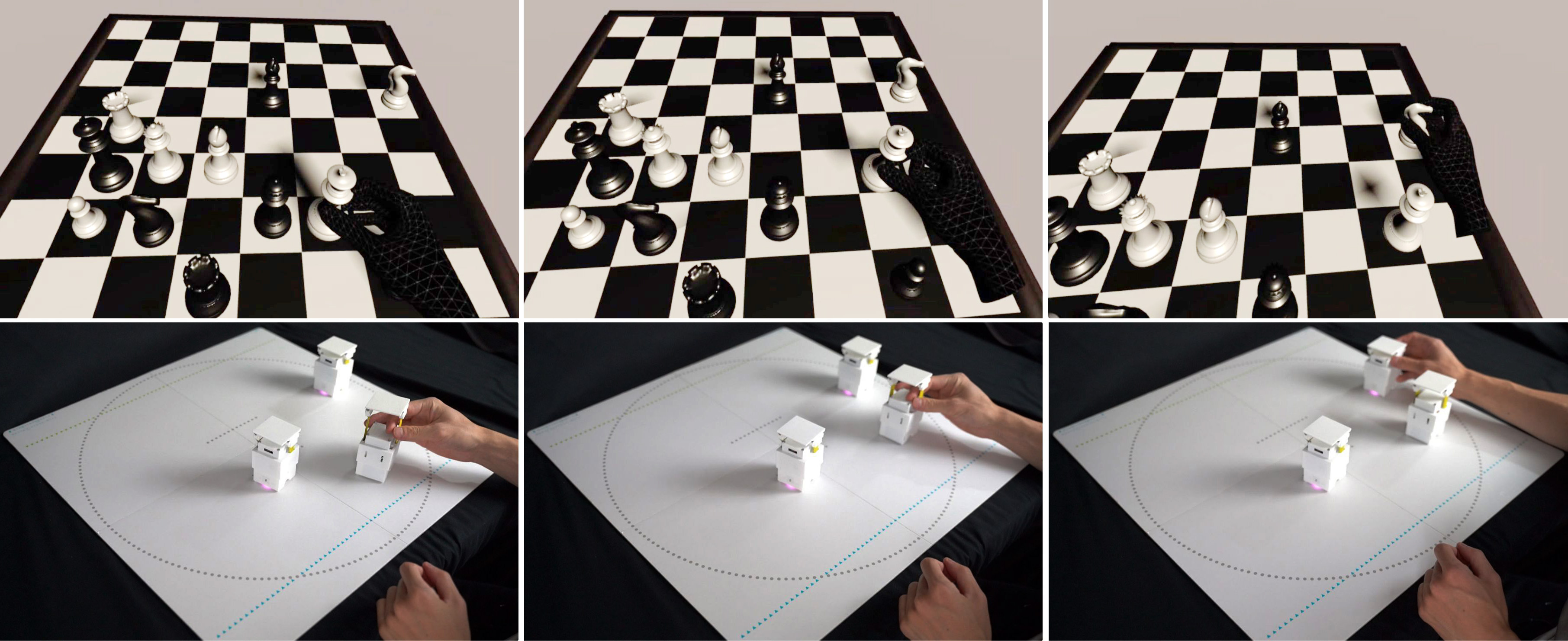}
\caption{Playing chess: When a user moves a physical piece, the virtual piece moves accordingly and is transmitted to the opponent.
}
\Description{
Modeling a virtual environment with HapticBots, shown in a sequence of three panels with the user’s virtual view, and below it a sequence of three panels showing the HapticBots and hands. The user touches 3D models and rearranges them by handling the robots.
}
~\label{fig:application-chess}
\end{figure}

Tangible interfaces can enrich remote collaboration through shared synchronized physical objects~\cite{brave1998tangible}. Using two connected \system{} setups, we can reproduce remote physical objects, or introduce shared virtual objects. Figure \ref{fig:application-chess} shows an example of a chess game application where the user moves the chess figures physically through robots. \eyal{As a user is replacing an opponent piece from the board, she can feel the robots at the correct place on the board. }
This interaction could extend to multiple end points to create shared, distributed multi-user spaces. 

Through its encountered-type haptic rendering approach, \system{} physically renders information about sizes, locations and heights of objects on-demand where the user touches them. \system{} also enables direct interaction with the 3D models, where users can pick up and move the robots to modify objects in the terrain and to redesign the environment (Figure \ref{fig:application-chess}).

\subsection{Gaming and Entertainment}

\begin{figure}[h!]
\centering
\includegraphics[width=1\columnwidth]{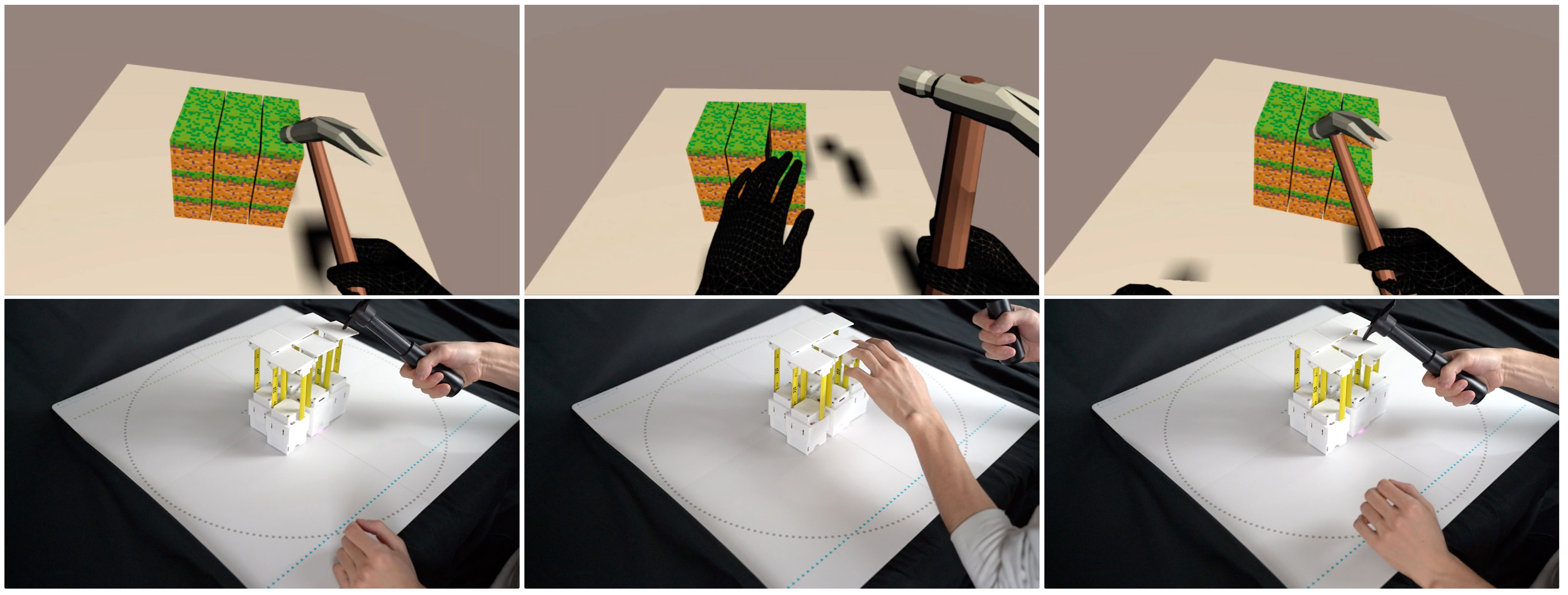}
\caption{Playing Minecraft: The robots render the virtual terrain. When the player hammers with a pickaxe on a real robot, it changes height and the virtual terrain is chipped away.}
\Description{
Playing Minecraft with HapticBots, shown in a sequence of three panels with the user’s virtual view, and below it a sequence of three panels showing the HapticBots and hands. When the player hammers with a pickaxe on a real robot, it changes height and the virtual terrain is chipped away.
}
~\label{fig:application-minecraft}
\end{figure}

World-building games like Minecraft often rely on players constructing terrains and objects. However the lack of haptics distracts from the immersive experience. \system{} can augment the game experience during construction or game play in these VR games. Apart from the previously mentioned interactions to grab, push, and encounter, multiple robots can act in coordinated ways to simulate larger objects. They can also provide proxy objects that interact with additional props and game controllers, such as an axe in Minecraft (see Figure~\ref{fig:application-minecraft}).

\section{Limitations and Future Work}

\system{} is our first functional prototype to demonstrate the concept of a distributed encountered-type haptic device. The following future work could improve this approach further.

\subsubsection*{Number of Robots and Coordinated Swarm Behaviors}
Our system employs relatively a small number of robots (e.g., 2-7 robots in the demonstrated applications).
This was partly because we noticed that the small number of robots are often sufficient to render various virtual objects in our interaction space (55 cm $\times$ 55 cm), as a single robot can also render the continuous surface.
However, by employing more robots, we could further leverage the swarm behavior of the robots and enrich the haptic interaction.
When scaling the robots, there are a couple of technical challenges that need to be solved.
For example, in our settings, the maximum number of Bluetooth pairing is limited with 7 bots on a single computer, but this can be scaled, for example, by using multiple Raspberry Pi as access points~\cite{nakagaki2020hermits}.
Path planning of the robots also introduces another challenge, as more sophisticated swarm controls would be required to avoid the collision.
On the other hand, the large number of robots could benefit to support more expressive haptic experiences as discussed in Section 3.
As future work, we will continuously explore interactions and applications that benefit from complex swarm behaviors.

\subsubsection*{Haptic Retargeting}
While our prototype is relatively fast and accurate, the current prototype still has some technical limitations, such as the speed of vertical actuation, the maximum height of the actuator, and speed and accuracy of locomotion, which causes generates the physical-virtual discrepancy.
For example, when the hand moves too fast, the robot cannot change its shape and/or position in time for a good fidelity haptic perception.
\diff{Also, our robots may need to retract or extend the actuator before traveling, which may also negatively impact the response time.}
To mitigate this mechanical limitation, we are interested in incorporating haptic retargeting~\cite{azmandian2016haptic}, specifically for the combination of continuous surface simulation which has not been explored in prior works~\cite{gonzalez2020reach}. 
Related to this, the better anticipation of potential touch events would also improve the system's performance. For example, we could leverage the user's behavior such as eye-gaze, or anticipate based on the context of the application.
We will further investigate how these approaches can enable a better experience.

\subsubsection*{Combining with Customizable Passive Haptic Proxy}
It is also interesting to see the possibilities to augment existing robots with custom haptic proxy.
For example, the Hermits system ~\cite{nakagaki2020hermits} introduced the ability for robots to reconfigure their functionality through mechanical shell add-ons. 
This way it is possible to augment the robot with various geometries (e.g., attached haptic props), surface textures (e.g., different robots have different materials), and even functional objects (e.g., buttons, sliders) that the user can feel when touching the robots.
Such ideas are partially explored in the prior works (e.g., adding a string for force feedback of fishing application~\cite{wang2020movevr}), but there should be broader design space for combining shape-changing mobile robots with external passive props. 
Also, the coordinated robots can actuate an existing environment, materials, and objects (e.g., similar to RoomShift~\cite{suzuki2020roomshift}, Sublimate~\cite{leithinger2013sublimate} or Molebot~\cite{lee2011molebot}).
We are interested in exploring the design space of this {\it inter-material interaction}~\cite{follmer2013inform} between \system{} and external environments.

\subsubsection*{User Force}
\diff{The stability of the robots is another technical limitation. 
We noticed that the robot’s movement becomes less stable---like the robot could be knocked over---while being touched, especially when the actuator is tilted or fully extended (e.g., over 25-30 cm).
This limitation can be alleviated by using a magnetic sheet and attaching a magnet to the robot.
}
Our current prototype also does not have the force-sensing capability, thus the input and interaction capability is limited.
An additional force sensing can expand the tangible interaction modality. 
For example, the user could push the actuator to lower the height or stretch it to extend. These interactions can bring additional rich embodied and tangible interactions into VR.
We are interested in further exploring this aspect.

\subsubsection*{Size and Resolution}
\diff{Finally, there is a trade-off between high resolution contact points and complexity of the position control.}
Currently, the footprint of our prototype robot is 5 $\times$ 5 cm. Scaling down to finger-tip size (e.g., 1-2 cm) would enable the higher resolution rendering and more portable system, but also introduces a number of technical challenges in actuation robustness.
For example, as we discussed in the implementation section, there is a trade-off between the size of the motor, the speed of actuation, and the active or holding force against the user's hand. The actuators' motor and gear torque also affect the stiffness of the tape---as the tape becomes stiffer, a stronger motor will be required to roll and unroll, which makes the overall size large.
\diff{However, unlike using swarm robots for shape-changing displays~\cite{suzuki2019shapebots}, the use for haptics in VR may not always require such a high-resolution contact point because we can leverage the visual illusion of the touch points.
Therefore, there might be a certain threshold of a minimum required size and a maximum number of robots.
Similar to visuo haptic illusion for shape displays~\cite{abtahi2018visuo}, investigating the scalability and trade-off of swarm robots for haptics in VR would be also an interesting research question in the future.
}

\section{Conclusion}
We introduced distributed encountered type haptics, a novel encountered type haptic concept with tabletop shape-changing robots.
Compared to previous encountered type haptics, such shape displays and robotic arm, our proposed approach improves the deployability and scalability of the system, while maintaining generalizability for general-purpose haptic applications.
This paper contributes to this concept, a set of haptic interaction techniques, and its demonstration with \system{} system.
The design and implementation of \system{} were described and the results of our user evaluation show the promise of our approach to simulating various shapes rendering.
Finally, we demonstrated various haptic applications for different VR scenarios, including training, gaming and entertainment, design, and remote collaboration.



\balance
\bibliographystyle{ACM-Reference-Format}
\bibliography{HBots_references}


\begin{thebibliography}{61}


\ifx \showCODEN    \undefined \def \showCODEN     #1{\unskip}     \fi
\ifx \showDOI      \undefined \def \showDOI       #1{#1}\fi
\ifx \showISBNx    \undefined \def \showISBNx     #1{\unskip}     \fi
\ifx \showISBNxiii \undefined \def \showISBNxiii  #1{\unskip}     \fi
\ifx \showISSN     \undefined \def \showISSN      #1{\unskip}     \fi
\ifx \showLCCN     \undefined \def \showLCCN      #1{\unskip}     \fi
\ifx \shownote     \undefined \def \shownote      #1{#1}          \fi
\ifx \showarticletitle \undefined \def \showarticletitle #1{#1}   \fi
\ifx \showURL      \undefined \def \showURL       {\relax}        \fi
\providecommand\bibfield[2]{#2}
\providecommand\bibinfo[2]{#2}
\providecommand\natexlab[1]{#1}
\providecommand\showeprint[2][]{arXiv:#2}

\bibitem[\protect\citeauthoryear{Abtahi and Follmer}{Abtahi and
  Follmer}{2018}]%
        {abtahi2018visuo}
\bibfield{author}{\bibinfo{person}{Parastoo Abtahi} {and} \bibinfo{person}{Sean
  Follmer}.} \bibinfo{year}{2018}\natexlab{}.
\newblock \showarticletitle{Visuo-haptic illusions for improving the perceived
  performance of shape displays}. In \bibinfo{booktitle}{\emph{Proceedings of
  the 2018 CHI Conference on Human Factors in Computing Systems}}.
  \bibinfo{pages}{1--13}.
\newblock


\bibitem[\protect\citeauthoryear{Abtahi, Landry, Yang, Pavone, Follmer, and
  Landay}{Abtahi et~al\mbox{.}}{2019}]%
        {abtahi2019beyond}
\bibfield{author}{\bibinfo{person}{Parastoo Abtahi}, \bibinfo{person}{Benoit
  Landry}, \bibinfo{person}{Jackie Yang}, \bibinfo{person}{Marco Pavone},
  \bibinfo{person}{Sean Follmer}, {and} \bibinfo{person}{James~A Landay}.}
  \bibinfo{year}{2019}\natexlab{}.
\newblock \showarticletitle{Beyond the force: Using quadcopters to appropriate
  objects and the environment for haptics in virtual reality}. In
  \bibinfo{booktitle}{\emph{Proceedings of the 2019 CHI Conference on Human
  Factors in Computing Systems}}. \bibinfo{pages}{1--13}.
\newblock


\bibitem[\protect\citeauthoryear{Araujo, Jota, Perumal, Yao, Singh, and
  Wigdor}{Araujo et~al\mbox{.}}{2016}]%
        {araujo2016snake}
\bibfield{author}{\bibinfo{person}{Bruno Araujo}, \bibinfo{person}{Ricardo
  Jota}, \bibinfo{person}{Varun Perumal}, \bibinfo{person}{Jia~Xian Yao},
  \bibinfo{person}{Karan Singh}, {and} \bibinfo{person}{Daniel Wigdor}.}
  \bibinfo{year}{2016}\natexlab{}.
\newblock \showarticletitle{Snake Charmer: Physically Enabling Virtual
  Objects}. In \bibinfo{booktitle}{\emph{Proceedings of the TEI'16: Tenth
  International Conference on Tangible, Embedded, and Embodied Interaction}}.
  ACM, \bibinfo{pages}{218--226}.
\newblock


\bibitem[\protect\citeauthoryear{Arora, Saini, Mehra, Jain, Shrey, and
  Parnami}{Arora et~al\mbox{.}}{2019}]%
        {arora2019virtualbricks}
\bibfield{author}{\bibinfo{person}{Jatin Arora}, \bibinfo{person}{Aryan Saini},
  \bibinfo{person}{Nirmita Mehra}, \bibinfo{person}{Varnit Jain},
  \bibinfo{person}{Shwetank Shrey}, {and} \bibinfo{person}{Aman Parnami}.}
  \bibinfo{year}{2019}\natexlab{}.
\newblock \showarticletitle{Virtualbricks: Exploring a scalable, modular
  toolkit for enabling physical manipulation in vr}. In
  \bibinfo{booktitle}{\emph{Proceedings of the 2019 CHI Conference on Human
  Factors in Computing Systems}}. \bibinfo{pages}{1--12}.
\newblock


\bibitem[\protect\citeauthoryear{Azmandian, Hancock, Benko, Ofek, and
  Wilson}{Azmandian et~al\mbox{.}}{2016}]%
        {azmandian2016haptic}
\bibfield{author}{\bibinfo{person}{Mahdi Azmandian}, \bibinfo{person}{Mark
  Hancock}, \bibinfo{person}{Hrvoje Benko}, \bibinfo{person}{Eyal Ofek}, {and}
  \bibinfo{person}{Andrew~D Wilson}.} \bibinfo{year}{2016}\natexlab{}.
\newblock \showarticletitle{Haptic retargeting: Dynamic repurposing of passive
  haptics for enhanced virtual reality experiences}. In
  \bibinfo{booktitle}{\emph{Proceedings of the 2016 chi conference on human
  factors in computing systems}}. ACM, \bibinfo{pages}{1968--1979}.
\newblock


\bibitem[\protect\citeauthoryear{Benko, Holz, Sinclair, and Ofek}{Benko
  et~al\mbox{.}}{2016}]%
        {benko2016normaltouch}
\bibfield{author}{\bibinfo{person}{Hrvoje Benko}, \bibinfo{person}{Christian
  Holz}, \bibinfo{person}{Mike Sinclair}, {and} \bibinfo{person}{Eyal Ofek}.}
  \bibinfo{year}{2016}\natexlab{}.
\newblock \showarticletitle{Normaltouch and texturetouch: High-fidelity 3d
  haptic shape rendering on handheld virtual reality controllers}. In
  \bibinfo{booktitle}{\emph{Proceedings of the 29th Annual Symposium on User
  Interface Software and Technology}}. ACM, \bibinfo{pages}{717--728}.
\newblock


\bibitem[\protect\citeauthoryear{Brave, Ishii, and Dahley}{Brave
  et~al\mbox{.}}{1998}]%
        {brave1998tangible}
\bibfield{author}{\bibinfo{person}{Scott Brave}, \bibinfo{person}{Hiroshi
  Ishii}, {and} \bibinfo{person}{Andrew Dahley}.}
  \bibinfo{year}{1998}\natexlab{}.
\newblock \showarticletitle{Tangible interfaces for remote collaboration and
  communication}. In \bibinfo{booktitle}{\emph{Proceedings of the 1998 ACM
  conference on Computer supported cooperative work}}.
  \bibinfo{pages}{169--178}.
\newblock


\bibitem[\protect\citeauthoryear{Cheng, Ofek, Holz, Benko, and Wilson}{Cheng
  et~al\mbox{.}}{2017}]%
        {cheng2017sparse}
\bibfield{author}{\bibinfo{person}{Lung-Pan Cheng}, \bibinfo{person}{Eyal
  Ofek}, \bibinfo{person}{Christian Holz}, \bibinfo{person}{Hrvoje Benko},
  {and} \bibinfo{person}{Andrew~D Wilson}.} \bibinfo{year}{2017}\natexlab{}.
\newblock \showarticletitle{Sparse haptic proxy: Touch feedback in virtual
  environments using a general passive prop}. In
  \bibinfo{booktitle}{\emph{Proceedings of the 2017 CHI Conference on Human
  Factors in Computing Systems}}. \bibinfo{pages}{3718--3728}.
\newblock


\bibitem[\protect\citeauthoryear{Choi, Hawkes, Christensen, Ploch, and
  Follmer}{Choi et~al\mbox{.}}{2016}]%
        {choi2016wolverine}
\bibfield{author}{\bibinfo{person}{Inrak Choi}, \bibinfo{person}{Elliot~W
  Hawkes}, \bibinfo{person}{David~L Christensen},
  \bibinfo{person}{Christopher~J Ploch}, {and} \bibinfo{person}{Sean Follmer}.}
  \bibinfo{year}{2016}\natexlab{}.
\newblock \showarticletitle{Wolverine: A wearable haptic interface for grasping
  in virtual reality}. In \bibinfo{booktitle}{\emph{2016 IEEE/RSJ International
  Conference on Intelligent Robots and Systems (IROS)}}. IEEE,
  \bibinfo{pages}{986--993}.
\newblock


\bibitem[\protect\citeauthoryear{Choi, Ofek, Benko, Sinclair, and Holz}{Choi
  et~al\mbox{.}}{2018}]%
        {choi2018claw}
\bibfield{author}{\bibinfo{person}{Inrak Choi}, \bibinfo{person}{Eyal Ofek},
  \bibinfo{person}{Hrvoje Benko}, \bibinfo{person}{Mike Sinclair}, {and}
  \bibinfo{person}{Christian Holz}.} \bibinfo{year}{2018}\natexlab{}.
\newblock \showarticletitle{Claw: A multifunctional handheld haptic controller
  for grasping, touching, and triggering in virtual reality}. In
  \bibinfo{booktitle}{\emph{Proceedings of the 2018 CHI Conference on Human
  Factors in Computing Systems}}. \bibinfo{pages}{1--13}.
\newblock


\bibitem[\protect\citeauthoryear{F{\aa}hraeus, Lindgren, and
  Burstr{\"o}m}{F{\aa}hraeus et~al\mbox{.}}{2007}]%
        {faahraeus2007electronic}
\bibfield{author}{\bibinfo{person}{Christer F{\aa}hraeus},
  \bibinfo{person}{Johan Lindgren}, {and} \bibinfo{person}{Stefan
  Burstr{\"o}m}.} \bibinfo{year}{2007}\natexlab{}.
\newblock \bibinfo{title}{Electronic pen}.
\newblock
\newblock
\newblock
\shownote{US Patent 7,239,306.}


\bibitem[\protect\citeauthoryear{Fang, Zhang, Dworman, and Harrison}{Fang
  et~al\mbox{.}}{2020}]%
        {wireality2020}
\bibfield{author}{\bibinfo{person}{Cathy Fang}, \bibinfo{person}{Yang Zhang},
  \bibinfo{person}{Matthew Dworman}, {and} \bibinfo{person}{Chris Harrison}.}
  \bibinfo{year}{2020}\natexlab{}.
\newblock \showarticletitle{Wireality: Enabling Complex Tangible Geometries in
  Virtual Reality with Worn Multi-String Haptics}. In
  \bibinfo{booktitle}{\emph{Proceedings of the 2020 CHI Conference on Human
  Factors in Computing Systems}} (Honolulu, HI, USA)
  \emph{(\bibinfo{series}{CHI '20})}. \bibinfo{publisher}{Association for
  Computing Machinery}, \bibinfo{address}{New York, NY, USA},
  \bibinfo{pages}{1–10}.
\newblock
\showISBNx{9781450367080}
\urldef\tempurl%
\url{https://doi.org/10.1145/3313831.3376470}
\showDOI{\tempurl}


\bibitem[\protect\citeauthoryear{Follmer, Leithinger, Olwal, Hogge, and
  Ishii}{Follmer et~al\mbox{.}}{2013}]%
        {follmer2013inform}
\bibfield{author}{\bibinfo{person}{Sean Follmer}, \bibinfo{person}{Daniel
  Leithinger}, \bibinfo{person}{Alex Olwal}, \bibinfo{person}{Akimitsu Hogge},
  {and} \bibinfo{person}{Hiroshi Ishii}.} \bibinfo{year}{2013}\natexlab{}.
\newblock \showarticletitle{inFORM: dynamic physical affordances and
  constraints through shape and object actuation.}. In
  \bibinfo{booktitle}{\emph{Uist}}, Vol.~\bibinfo{volume}{13}.
  \bibinfo{pages}{2501988--2502032}.
\newblock


\bibitem[\protect\citeauthoryear{Gonzalez, Abtahi, and Follmer}{Gonzalez
  et~al\mbox{.}}{2020}]%
        {gonzalez2020reach}
\bibfield{author}{\bibinfo{person}{Eric~J Gonzalez}, \bibinfo{person}{Parastoo
  Abtahi}, {and} \bibinfo{person}{Sean Follmer}.}
  \bibinfo{year}{2020}\natexlab{}.
\newblock \showarticletitle{REACH+: Extending the Reachability of
  Encountered-type Haptics Devices through Dynamic Redirection in VR}. In
  \bibinfo{booktitle}{\emph{Proceedings of the 33rd Annual ACM Symposium on
  User Interface Software and Technology}}. ACM.
\newblock


\bibitem[\protect\citeauthoryear{Guinness, Muehlbradt, Szafir, and
  Kane}{Guinness et~al\mbox{.}}{2019}]%
        {guinness2019robographics}
\bibfield{author}{\bibinfo{person}{Darren Guinness}, \bibinfo{person}{Annika
  Muehlbradt}, \bibinfo{person}{Daniel Szafir}, {and} \bibinfo{person}{Shaun~K
  Kane}.} \bibinfo{year}{2019}\natexlab{}.
\newblock \showarticletitle{RoboGraphics: Dynamic Tactile Graphics Powered by
  Mobile Robots}. In \bibinfo{booktitle}{\emph{The 21st International ACM
  SIGACCESS Conference on Computers and Accessibility}}.
  \bibinfo{pages}{318--328}.
\newblock


\bibitem[\protect\citeauthoryear{Hammond, Usevitch, Hawkes, and
  Follmer}{Hammond et~al\mbox{.}}{2017}]%
        {hammond2017pneumatic}
\bibfield{author}{\bibinfo{person}{Zachary~M Hammond},
  \bibinfo{person}{Nathan~S Usevitch}, \bibinfo{person}{Elliot~W Hawkes}, {and}
  \bibinfo{person}{Sean Follmer}.} \bibinfo{year}{2017}\natexlab{}.
\newblock \showarticletitle{Pneumatic reel actuator: Design, modeling, and
  implementation}. In \bibinfo{booktitle}{\emph{Robotics and Automation (ICRA),
  2017 IEEE International Conference on}}. IEEE, \bibinfo{pages}{626--633}.
\newblock


\bibitem[\protect\citeauthoryear{Hayward, Astley, Cruz-Hernandez, Grant, and
  Robles-De-La-Torre}{Hayward et~al\mbox{.}}{2004}]%
        {hayward2004haptic}
\bibfield{author}{\bibinfo{person}{Vincent Hayward}, \bibinfo{person}{Oliver~R
  Astley}, \bibinfo{person}{Manuel Cruz-Hernandez}, \bibinfo{person}{Danny
  Grant}, {and} \bibinfo{person}{Gabriel Robles-De-La-Torre}.}
  \bibinfo{year}{2004}\natexlab{}.
\newblock \showarticletitle{Haptic interfaces and devices}.
\newblock \bibinfo{journal}{\emph{Sensor review}} (\bibinfo{year}{2004}).
\newblock


\bibitem[\protect\citeauthoryear{He, Zhu, and Perlin}{He et~al\mbox{.}}{2017}]%
        {he2017physhare}
\bibfield{author}{\bibinfo{person}{Zhenyi He}, \bibinfo{person}{Fengyuan Zhu},
  {and} \bibinfo{person}{Ken Perlin}.} \bibinfo{year}{2017}\natexlab{}.
\newblock \showarticletitle{Physhare: sharing physical interaction in virtual
  reality}. In \bibinfo{booktitle}{\emph{Proceedings of the 30th Annual ACM
  Symposium on User Interface Software and Technology}}. ACM,
  \bibinfo{pages}{17--19}.
\newblock


\bibitem[\protect\citeauthoryear{Heo, Chung, Lee, and Wigdor}{Heo
  et~al\mbox{.}}{2018}]%
        {heo2018thor}
\bibfield{author}{\bibinfo{person}{Seongkook Heo}, \bibinfo{person}{Christina
  Chung}, \bibinfo{person}{Geehyuk Lee}, {and} \bibinfo{person}{Daniel
  Wigdor}.} \bibinfo{year}{2018}\natexlab{}.
\newblock \showarticletitle{Thor's hammer: An ungrounded force feedback device
  utilizing propeller-induced propulsive force}. In
  \bibinfo{booktitle}{\emph{Proceedings of the 2018 CHI Conference on Human
  Factors in Computing Systems}}. \bibinfo{pages}{1--11}.
\newblock


\bibitem[\protect\citeauthoryear{Hettiarachchi and Wigdor}{Hettiarachchi and
  Wigdor}{2016}]%
        {hettiarachchi2016annexing}
\bibfield{author}{\bibinfo{person}{Anuruddha Hettiarachchi} {and}
  \bibinfo{person}{Daniel Wigdor}.} \bibinfo{year}{2016}\natexlab{}.
\newblock \showarticletitle{Annexing reality: Enabling opportunistic use of
  everyday objects as tangible proxies in augmented reality}. In
  \bibinfo{booktitle}{\emph{Proceedings of the 2016 CHI Conference on Human
  Factors in Computing Systems}}. ACM, \bibinfo{pages}{1957--1967}.
\newblock


\bibitem[\protect\citeauthoryear{Hirota and Hirose}{Hirota and Hirose}{1995}]%
        {hirota1995simulation}
\bibfield{author}{\bibinfo{person}{Koichi Hirota} {and}
  \bibinfo{person}{Michitaka Hirose}.} \bibinfo{year}{1995}\natexlab{}.
\newblock \showarticletitle{Simulation and presentation of curved surface in
  virtual reality environment through surface display}. In
  \bibinfo{booktitle}{\emph{Proceedings Virtual Reality Annual International
  Symposium'95}}. IEEE, \bibinfo{pages}{211--216}.
\newblock


\bibitem[\protect\citeauthoryear{Hoppe, Knierim, Kosch, Funk, Futami,
  Schneegass, Henze, Schmidt, and Machulla}{Hoppe et~al\mbox{.}}{2018}]%
        {hoppe2018vrhapticdrones}
\bibfield{author}{\bibinfo{person}{Matthias Hoppe}, \bibinfo{person}{Pascal
  Knierim}, \bibinfo{person}{Thomas Kosch}, \bibinfo{person}{Markus Funk},
  \bibinfo{person}{Lauren Futami}, \bibinfo{person}{Stefan Schneegass},
  \bibinfo{person}{Niels Henze}, \bibinfo{person}{Albrecht Schmidt}, {and}
  \bibinfo{person}{Tonja Machulla}.} \bibinfo{year}{2018}\natexlab{}.
\newblock \showarticletitle{VRHapticDrones: Providing Haptics in Virtual
  Reality through Quadcopters}. In \bibinfo{booktitle}{\emph{Proceedings of the
  17th International Conference on Mobile and Ubiquitous Multimedia}}. ACM,
  \bibinfo{pages}{7--18}.
\newblock


\bibitem[\protect\citeauthoryear{Insko, Meehan, Whitton, and Brooks}{Insko
  et~al\mbox{.}}{2001}]%
        {insko2001passive}
\bibfield{author}{\bibinfo{person}{Brent~Edward Insko}, \bibinfo{person}{M
  Meehan}, \bibinfo{person}{M Whitton}, {and} \bibinfo{person}{F Brooks}.}
  \bibinfo{year}{2001}\natexlab{}.
\newblock \emph{\bibinfo{title}{Passive haptics significantly enhances virtual
  environments}}.
\newblock \bibinfo{thesistype}{Ph.D. Dissertation}. \bibinfo{school}{University
  of North Carolina at Chapel Hill}.
\newblock


\bibitem[\protect\citeauthoryear{Iwata}{Iwata}{1999}]%
        {iwata1999walking}
\bibfield{author}{\bibinfo{person}{Hiroo Iwata}.}
  \bibinfo{year}{1999}\natexlab{}.
\newblock \showarticletitle{Walking about virtual environments on an infinite
  floor}. In \bibinfo{booktitle}{\emph{Proceedings IEEE Virtual Reality (Cat.
  No. 99CB36316)}}. IEEE, \bibinfo{pages}{286--293}.
\newblock


\bibitem[\protect\citeauthoryear{Iwata, Yano, Nakaizumi, and Kawamura}{Iwata
  et~al\mbox{.}}{2001}]%
        {iwata2001project}
\bibfield{author}{\bibinfo{person}{Hiroo Iwata}, \bibinfo{person}{Hiroaki
  Yano}, \bibinfo{person}{Fumitaka Nakaizumi}, {and} \bibinfo{person}{Ryo
  Kawamura}.} \bibinfo{year}{2001}\natexlab{}.
\newblock \showarticletitle{Project FEELEX: adding haptic surface to graphics}.
  In \bibinfo{booktitle}{\emph{Proceedings of the 28th annual conference on
  Computer graphics and interactive techniques}}. ACM,
  \bibinfo{pages}{469--476}.
\newblock


\bibitem[\protect\citeauthoryear{Kim and Nam}{Kim and Nam}{2015}]%
        {kim2015g}
\bibfield{author}{\bibinfo{person}{Chang-Min Kim} {and}
  \bibinfo{person}{Tek-Jin Nam}.} \bibinfo{year}{2015}\natexlab{}.
\newblock \showarticletitle{G-raff: an elevating tangible block for spatial
  tabletop interaction}. In \bibinfo{booktitle}{\emph{Proceedings of the 33rd
  Annual ACM Conference on Human Factors in Computing Systems}}.
  \bibinfo{pages}{4161--4164}.
\newblock


\bibitem[\protect\citeauthoryear{Kim and Follmer}{Kim and Follmer}{2017}]%
        {kim2017ubiswarm}
\bibfield{author}{\bibinfo{person}{Lawrence~H Kim} {and} \bibinfo{person}{Sean
  Follmer}.} \bibinfo{year}{2017}\natexlab{}.
\newblock \showarticletitle{Ubiswarm: Ubiquitous robotic interfaces and
  investigation of abstract motion as a display}.
\newblock \bibinfo{journal}{\emph{Proceedings of the ACM on Interactive,
  Mobile, Wearable and Ubiquitous Technologies}} \bibinfo{volume}{1},
  \bibinfo{number}{3} (\bibinfo{year}{2017}), \bibinfo{pages}{66}.
\newblock


\bibitem[\protect\citeauthoryear{Kim and Follmer}{Kim and Follmer}{2019}]%
        {kim2019swarmhaptics}
\bibfield{author}{\bibinfo{person}{Lawrence~H Kim} {and} \bibinfo{person}{Sean
  Follmer}.} \bibinfo{year}{2019}\natexlab{}.
\newblock \showarticletitle{Swarmhaptics: Haptic display with swarm robots}. In
  \bibinfo{booktitle}{\emph{Proceedings of the 2019 CHI conference on human
  factors in computing systems}}. \bibinfo{pages}{1--13}.
\newblock


\bibitem[\protect\citeauthoryear{Kohli, Burns, Miller, and Fuchs}{Kohli
  et~al\mbox{.}}{2005}]%
        {kohli2005combining}
\bibfield{author}{\bibinfo{person}{Luv Kohli}, \bibinfo{person}{Eric Burns},
  \bibinfo{person}{Dorian Miller}, {and} \bibinfo{person}{Henry Fuchs}.}
  \bibinfo{year}{2005}\natexlab{}.
\newblock \showarticletitle{Combining passive haptics with redirected walking}.
  In \bibinfo{booktitle}{\emph{Proceedings of the 2005 international conference
  on Augmented tele-existence}}. ACM, \bibinfo{pages}{253--254}.
\newblock


\bibitem[\protect\citeauthoryear{Kovacs, Ofek, Gonzalez-Franco, Siu, Marwecki,
  Holz, and Sinclair}{Kovacs et~al\mbox{.}}{2020}]%
        {Kovacs2020HapticPivot}
\bibfield{author}{\bibinfo{person}{Robert Kovacs}, \bibinfo{person}{Eyal Ofek},
  \bibinfo{person}{Mar Gonzalez-Franco}, \bibinfo{person}{Alexa~F Siu},
  \bibinfo{person}{Sebastian Marwecki}, \bibinfo{person}{Christian Holz}, {and}
  \bibinfo{person}{Mike Sinclair}.} \bibinfo{year}{2020}\natexlab{}.
\newblock \showarticletitle{Haptic PIVOT: On-Demand Handhelds in VR}. In
  \bibinfo{booktitle}{\emph{Proceedings of the 33nd Annual ACM Symposium on
  User Interface Software and Technology}} \emph{(\bibinfo{series}{UIST '20})}.
  \bibinfo{publisher}{Association for Computing Machinery},
  \bibinfo{address}{New York, NY, USA}.
\newblock
\showISBNx{9781450368162}


\bibitem[\protect\citeauthoryear{Le~Goc, Kim, Parsaei, Fekete, Dragicevic, and
  Follmer}{Le~Goc et~al\mbox{.}}{2016}]%
        {le2016zooids}
\bibfield{author}{\bibinfo{person}{Mathieu Le~Goc}, \bibinfo{person}{Lawrence~H
  Kim}, \bibinfo{person}{Ali Parsaei}, \bibinfo{person}{Jean-Daniel Fekete},
  \bibinfo{person}{Pierre Dragicevic}, {and} \bibinfo{person}{Sean Follmer}.}
  \bibinfo{year}{2016}\natexlab{}.
\newblock \showarticletitle{Zooids: Building blocks for swarm user interfaces}.
  In \bibinfo{booktitle}{\emph{Proceedings of the 29th Annual Symposium on User
  Interface Software and Technology}}. ACM, \bibinfo{pages}{97--109}.
\newblock


\bibitem[\protect\citeauthoryear{Lee, Sinclair, Gonzalez-Franco, Ofek, and
  Holz}{Lee et~al\mbox{.}}{2019}]%
        {lee2019torc}
\bibfield{author}{\bibinfo{person}{Jaeyeon Lee}, \bibinfo{person}{Mike
  Sinclair}, \bibinfo{person}{Mar Gonzalez-Franco}, \bibinfo{person}{Eyal
  Ofek}, {and} \bibinfo{person}{Christian Holz}.}
  \bibinfo{year}{2019}\natexlab{}.
\newblock \showarticletitle{TORC: A virtual reality controller for in-hand
  high-dexterity finger interaction}. In \bibinfo{booktitle}{\emph{Proceedings
  of the 2019 CHI Conference on Human Factors in Computing Systems}}.
  \bibinfo{pages}{1--13}.
\newblock


\bibitem[\protect\citeauthoryear{Lee, Kim, Lee, Shin, and Lee}{Lee
  et~al\mbox{.}}{2011}]%
        {lee2011molebot}
\bibfield{author}{\bibinfo{person}{Narae Lee}, \bibinfo{person}{Juwhan Kim},
  \bibinfo{person}{Jungsoo Lee}, \bibinfo{person}{Myeongsoo Shin}, {and}
  \bibinfo{person}{Woohun Lee}.} \bibinfo{year}{2011}\natexlab{}.
\newblock \showarticletitle{Molebot: mole in a table}.
\newblock In \bibinfo{booktitle}{\emph{ACM SIGGRAPH 2011 Emerging
  Technologies}}. \bibinfo{pages}{1--1}.
\newblock


\bibitem[\protect\citeauthoryear{Leithinger, Follmer, Olwal, and
  Ishii}{Leithinger et~al\mbox{.}}{2015}]%
        {leithinger2015shape}
\bibfield{author}{\bibinfo{person}{Daniel Leithinger}, \bibinfo{person}{Sean
  Follmer}, \bibinfo{person}{Alex Olwal}, {and} \bibinfo{person}{Hiroshi
  Ishii}.} \bibinfo{year}{2015}\natexlab{}.
\newblock \showarticletitle{Shape displays: Spatial interaction with dynamic
  physical form}.
\newblock \bibinfo{journal}{\emph{IEEE computer graphics and applications}}
  \bibinfo{volume}{35}, \bibinfo{number}{5} (\bibinfo{year}{2015}),
  \bibinfo{pages}{5--11}.
\newblock


\bibitem[\protect\citeauthoryear{Leithinger, Follmer, Olwal, Luescher, Hogge,
  Lee, and Ishii}{Leithinger et~al\mbox{.}}{2013}]%
        {leithinger2013sublimate}
\bibfield{author}{\bibinfo{person}{Daniel Leithinger}, \bibinfo{person}{Sean
  Follmer}, \bibinfo{person}{Alex Olwal}, \bibinfo{person}{Samuel Luescher},
  \bibinfo{person}{Akimitsu Hogge}, \bibinfo{person}{Jinha Lee}, {and}
  \bibinfo{person}{Hiroshi Ishii}.} \bibinfo{year}{2013}\natexlab{}.
\newblock \showarticletitle{Sublimate: state-changing virtual and physical
  rendering to augment interaction with shape displays}. In
  \bibinfo{booktitle}{\emph{Proceedings of the SIGCHI conference on human
  factors in computing systems}}. \bibinfo{pages}{1441--1450}.
\newblock


\bibitem[\protect\citeauthoryear{McNeely}{McNeely}{1993}]%
        {mcneely1993robotic}
\bibfield{author}{\bibinfo{person}{William~A McNeely}.}
  \bibinfo{year}{1993}\natexlab{}.
\newblock \showarticletitle{Robotic graphics: a new approach to force feedback
  for virtual reality}. In \bibinfo{booktitle}{\emph{Proceedings of IEEE
  Virtual Reality Annual International Symposium}}. IEEE,
  \bibinfo{pages}{336--341}.
\newblock


\bibitem[\protect\citeauthoryear{Nakagaki, Fitzgerald, Ma, Vink, Levine, and
  Ishii}{Nakagaki et~al\mbox{.}}{2019}]%
        {nakagaki2019inforce}
\bibfield{author}{\bibinfo{person}{Ken Nakagaki}, \bibinfo{person}{Daniel
  Fitzgerald}, \bibinfo{person}{Zhiyao Ma}, \bibinfo{person}{Luke Vink},
  \bibinfo{person}{Daniel Levine}, {and} \bibinfo{person}{Hiroshi Ishii}.}
  \bibinfo{year}{2019}\natexlab{}.
\newblock \showarticletitle{inFORCE: Bi-directionalForce'Shape Display for
  Haptic Interaction}. In \bibinfo{booktitle}{\emph{Proceedings of the
  Thirteenth International Conference on Tangible, Embedded, and Embodied
  Interaction}}. \bibinfo{pages}{615--623}.
\newblock


\bibitem[\protect\citeauthoryear{Nakagaki, Leong, Tappa, Wilbert, and
  Ishii}{Nakagaki et~al\mbox{.}}{2020}]%
        {nakagaki2020hermits}
\bibfield{author}{\bibinfo{person}{Ken Nakagaki}, \bibinfo{person}{Joanne
  Leong}, \bibinfo{person}{Jordan~L Tappa}, \bibinfo{person}{Joao Wilbert},
  {and} \bibinfo{person}{Hiroshi Ishii}.} \bibinfo{year}{2020}\natexlab{}.
\newblock \showarticletitle{HERMITS: Dynamically Reconfiguring the
  Interactivity of Self-Propelled TUIs with Mechanical Shell Add-ons}. In
  \bibinfo{booktitle}{\emph{Proceedings of the 33rd Annual ACM Symposium on
  User Interface Software and Technology}}. ACM.
\newblock


\bibitem[\protect\citeauthoryear{Patten and Ishii}{Patten and Ishii}{2007}]%
        {patten2007mechanical}
\bibfield{author}{\bibinfo{person}{James Patten} {and} \bibinfo{person}{Hiroshi
  Ishii}.} \bibinfo{year}{2007}\natexlab{}.
\newblock \showarticletitle{Mechanical constraints as computational constraints
  in tabletop tangible interfaces}. In \bibinfo{booktitle}{\emph{Proceedings of
  the SIGCHI conference on Human factors in computing systems}}.
  \bibinfo{pages}{809--818}.
\newblock


\bibitem[\protect\citeauthoryear{Razzaque, Kohn, and Whitton}{Razzaque
  et~al\mbox{.}}{2005}]%
        {razzaque2005redirected}
\bibfield{author}{\bibinfo{person}{Sharif Razzaque}, \bibinfo{person}{Zachariah
  Kohn}, {and} \bibinfo{person}{Mary~C Whitton}.}
  \bibinfo{year}{2005}\natexlab{}.
\newblock \bibinfo{booktitle}{\emph{Redirected walking}}.
\newblock \bibinfo{publisher}{Citeseer}.
\newblock


\bibitem[\protect\citeauthoryear{Simeone, Velloso, and Gellersen}{Simeone
  et~al\mbox{.}}{2015}]%
        {simeone2015substitutional}
\bibfield{author}{\bibinfo{person}{Adalberto~L Simeone},
  \bibinfo{person}{Eduardo Velloso}, {and} \bibinfo{person}{Hans Gellersen}.}
  \bibinfo{year}{2015}\natexlab{}.
\newblock \showarticletitle{Substitutional reality: Using the physical
  environment to design virtual reality experiences}. In
  \bibinfo{booktitle}{\emph{Proceedings of the 33rd Annual ACM Conference on
  Human Factors in Computing Systems}}. \bibinfo{pages}{3307--3316}.
\newblock


\bibitem[\protect\citeauthoryear{Sinclair, Ofek, Gonzalez-Franco, and
  Holz}{Sinclair et~al\mbox{.}}{2019}]%
        {Sinclair2019Capstan}
\bibfield{author}{\bibinfo{person}{Mike Sinclair}, \bibinfo{person}{Eyal Ofek},
  \bibinfo{person}{Mar Gonzalez-Franco}, {and} \bibinfo{person}{Christian
  Holz}.} \bibinfo{year}{2019}\natexlab{}.
\newblock \showarticletitle{CapstanCrunch: A Haptic VR Controller with
  User-Supplied Force Feedback}. In \bibinfo{booktitle}{\emph{Proceedings of
  the 32nd Annual ACM Symposium on User Interface Software and Technology}}
  (New Orleans, LA, USA) \emph{(\bibinfo{series}{UIST '19})}.
  \bibinfo{publisher}{Association for Computing Machinery},
  \bibinfo{address}{New York, NY, USA}, \bibinfo{pages}{815–829}.
\newblock
\showISBNx{9781450368162}
\urldef\tempurl%
\url{https://doi.org/10.1145/3332165.3347891}
\showDOI{\tempurl}


\bibitem[\protect\citeauthoryear{Siu, Gonzalez, Yuan, Ginsberg, and
  Follmer}{Siu et~al\mbox{.}}{2018}]%
        {siu2018shapeshift}
\bibfield{author}{\bibinfo{person}{Alexa~F Siu}, \bibinfo{person}{Eric~J
  Gonzalez}, \bibinfo{person}{Shenli Yuan}, \bibinfo{person}{Jason~B Ginsberg},
  {and} \bibinfo{person}{Sean Follmer}.} \bibinfo{year}{2018}\natexlab{}.
\newblock \showarticletitle{Shapeshift: 2D spatial manipulation and
  self-actuation of tabletop shape displays for tangible and haptic
  interaction}. In \bibinfo{booktitle}{\emph{Proceedings of the 2018 CHI
  Conference on Human Factors in Computing Systems}}. ACM,
  \bibinfo{pages}{291}.
\newblock


\bibitem[\protect\citeauthoryear{Steed, Ofek, Sinclair, and
  Gonzalez-Franco}{Steed et~al\mbox{.}}{2021}]%
        {steed2021mechatronic}
\bibfield{author}{\bibinfo{person}{Anthony Steed}, \bibinfo{person}{Eyal Ofek},
  \bibinfo{person}{Mike Sinclair}, {and} \bibinfo{person}{Mar
  Gonzalez-Franco}.} \bibinfo{year}{2021}\natexlab{}.
\newblock \showarticletitle{A Mechatronic Shape Display based on Auxetic
  Materials}. In \bibinfo{booktitle}{\emph{Nature Communications}}.
\newblock


\bibitem[\protect\citeauthoryear{Sun, Yoshida, Narumi, and Hirose}{Sun
  et~al\mbox{.}}{2019}]%
        {sun2019PoCoPo}
\bibfield{author}{\bibinfo{person}{Yuqian Sun}, \bibinfo{person}{Shigeo
  Yoshida}, \bibinfo{person}{Takuji Narumi}, {and} \bibinfo{person}{Michitaka
  Hirose}.} \bibinfo{year}{2019}\natexlab{}.
\newblock \showarticletitle{PoCoPo: A handheld vr device for rendering size,
  shape, and stiffness of virtual objects in tool-based interactions}. In
  \bibinfo{booktitle}{\emph{Proceedings of the 2019 CHI Conference on Human
  Factors in Computing Systems}}. \bibinfo{pages}{1--12}.
\newblock


\bibitem[\protect\citeauthoryear{Suzuki, Hedayati, Zheng, Bohn, Szafir, Do,
  Gross, and Leithinger}{Suzuki et~al\mbox{.}}{2020a}]%
        {suzuki2020roomshift}
\bibfield{author}{\bibinfo{person}{Ryo Suzuki}, \bibinfo{person}{Hooman
  Hedayati}, \bibinfo{person}{Clement Zheng}, \bibinfo{person}{James~L Bohn},
  \bibinfo{person}{Daniel Szafir}, \bibinfo{person}{Ellen Yi-Luen Do},
  \bibinfo{person}{Mark~D Gross}, {and} \bibinfo{person}{Daniel Leithinger}.}
  \bibinfo{year}{2020}\natexlab{a}.
\newblock \showarticletitle{RoomShift: Room-scale Dynamic Haptics for VR with
  Furniture-moving Swarm Robots}. In \bibinfo{booktitle}{\emph{Proceedings of
  the 2020 CHI Conference on Human Factors in Computing Systems}}.
  \bibinfo{pages}{1--11}.
\newblock


\bibitem[\protect\citeauthoryear{Suzuki, Kato, Gross, and Yeh}{Suzuki
  et~al\mbox{.}}{2018}]%
        {suzuki2018reactile}
\bibfield{author}{\bibinfo{person}{Ryo Suzuki}, \bibinfo{person}{Jun Kato},
  \bibinfo{person}{Mark~D Gross}, {and} \bibinfo{person}{Tom Yeh}.}
  \bibinfo{year}{2018}\natexlab{}.
\newblock \showarticletitle{Reactile: Programming Swarm User Interfaces through
  Direct Physical Manipulation}. In \bibinfo{booktitle}{\emph{Proceedings of
  the 2018 CHI Conference on Human Factors in Computing Systems}}. ACM,
  \bibinfo{pages}{199}.
\newblock


\bibitem[\protect\citeauthoryear{Suzuki, Nakayama, Liu, Kakehi, Gross, and
  Leithinger}{Suzuki et~al\mbox{.}}{2020b}]%
        {suzuki2020lifttiles}
\bibfield{author}{\bibinfo{person}{Ryo Suzuki}, \bibinfo{person}{Ryosuke
  Nakayama}, \bibinfo{person}{Dan Liu}, \bibinfo{person}{Yasuaki Kakehi},
  \bibinfo{person}{Mark~D. Gross}, {and} \bibinfo{person}{Daniel Leithinger}.}
  \bibinfo{year}{2020}\natexlab{b}.
\newblock \showarticletitle{LiftTiles: Constructive Building Blocks for
  Prototyping Room-scale Shape-changing Interfaces}. In
  \bibinfo{booktitle}{\emph{Proceedings of the Fourteenth International
  Conference on Tangible, Embedded, and Embodied Interaction}}. ACM.
\newblock


\bibitem[\protect\citeauthoryear{Suzuki, Stangl, Gross, and Yeh}{Suzuki
  et~al\mbox{.}}{2017}]%
        {suzuki2017fluxmarker}
\bibfield{author}{\bibinfo{person}{Ryo Suzuki}, \bibinfo{person}{Abigale
  Stangl}, \bibinfo{person}{Mark~D Gross}, {and} \bibinfo{person}{Tom Yeh}.}
  \bibinfo{year}{2017}\natexlab{}.
\newblock \showarticletitle{FluxMarker: Enhancing Tactile Graphics with Dynamic
  Tactile Markers}. In \bibinfo{booktitle}{\emph{Proceedings of the 19th
  International ACM SIGACCESS Conference on Computers and Accessibility}}.
  \bibinfo{pages}{190--199}.
\newblock


\bibitem[\protect\citeauthoryear{Suzuki, Zheng, Yeh, Yi-Luen~Do, Gross, and
  Leithinger}{Suzuki et~al\mbox{.}}{2019}]%
        {suzuki2019shapebots}
\bibfield{author}{\bibinfo{person}{Ryo Suzuki}, \bibinfo{person}{Yasuaki Zheng,
  Clement~Kakehi}, \bibinfo{person}{Tom Yeh}, \bibinfo{person}{Ellen
  Yi-Luen~Do}, \bibinfo{person}{Mark~D Gross}, {and} \bibinfo{person}{Daniel
  Leithinger}.} \bibinfo{year}{2019}\natexlab{}.
\newblock \showarticletitle{ShapeBots: Shape-changing Swarm Robots}. In
  \bibinfo{booktitle}{\emph{Proceedings of the 32nd Annual ACM Symposium on
  User Interface Software and Technology}}. ACM.
\newblock


\bibitem[\protect\citeauthoryear{Takei, Iida, and Naemura}{Takei
  et~al\mbox{.}}{2011}]%
        {takei2011kinereels}
\bibfield{author}{\bibinfo{person}{Shohei Takei}, \bibinfo{person}{Makoto
  Iida}, {and} \bibinfo{person}{Takeshi Naemura}.}
  \bibinfo{year}{2011}\natexlab{}.
\newblock \showarticletitle{Kinereels: extension actuators for dynamic 3d
  shape}. In \bibinfo{booktitle}{\emph{ACM SIGGRAPH 2011 Posters}}. ACM,
  \bibinfo{pages}{84}.
\newblock


\bibitem[\protect\citeauthoryear{Teng, Kuo, Wang, Chiang, Huang, Chan, and
  Chen}{Teng et~al\mbox{.}}{2018}]%
        {teng2018pupop}
\bibfield{author}{\bibinfo{person}{Shan-Yuan Teng}, \bibinfo{person}{Tzu-Sheng
  Kuo}, \bibinfo{person}{Chi Wang}, \bibinfo{person}{Chi-huan Chiang},
  \bibinfo{person}{Da-Yuan Huang}, \bibinfo{person}{Liwei Chan}, {and}
  \bibinfo{person}{Bing-Yu Chen}.} \bibinfo{year}{2018}\natexlab{}.
\newblock \showarticletitle{Pupop: Pop-up prop on palm for virtual reality}. In
  \bibinfo{booktitle}{\emph{Proceedings of the 31st Annual ACM Symposium on
  User Interface Software and Technology}}. \bibinfo{pages}{5--17}.
\newblock


\bibitem[\protect\citeauthoryear{Teng, Lin, Chiang, Kuo, Chan, Huang, and
  Chen}{Teng et~al\mbox{.}}{2019}]%
        {teng2019tilepop}
\bibfield{author}{\bibinfo{person}{Shan-Yuan Teng}, \bibinfo{person}{Cheng-Lung
  Lin}, \bibinfo{person}{Chi-huan Chiang}, \bibinfo{person}{Tzu-Sheng Kuo},
  \bibinfo{person}{Liwei Chan}, \bibinfo{person}{Da-Yuan Huang}, {and}
  \bibinfo{person}{Bing-Yu Chen}.} \bibinfo{year}{2019}\natexlab{}.
\newblock \showarticletitle{TilePoP: Tile-type Pop-up Prop for Virtual
  Reality}. In \bibinfo{booktitle}{\emph{Proceedings of the 32nd Annual ACM
  Symposium on User Interface Software and Technology}}. ACM.
\newblock


\bibitem[\protect\citeauthoryear{Van~den Berg, Lin, and Manocha}{Van~den Berg
  et~al\mbox{.}}{2008}]%
        {van2008reciprocal}
\bibfield{author}{\bibinfo{person}{Jur Van~den Berg}, \bibinfo{person}{Ming
  Lin}, {and} \bibinfo{person}{Dinesh Manocha}.}
  \bibinfo{year}{2008}\natexlab{}.
\newblock \showarticletitle{Reciprocal velocity obstacles for real-time
  multi-agent navigation}. In \bibinfo{booktitle}{\emph{2008 IEEE International
  Conference on Robotics and Automation}}. IEEE, \bibinfo{pages}{1928--1935}.
\newblock


\bibitem[\protect\citeauthoryear{Vonach, Gatterer, and Kaufmann}{Vonach
  et~al\mbox{.}}{2017}]%
        {vonach2017vrrobot}
\bibfield{author}{\bibinfo{person}{Emanuel Vonach}, \bibinfo{person}{Clemens
  Gatterer}, {and} \bibinfo{person}{Hannes Kaufmann}.}
  \bibinfo{year}{2017}\natexlab{}.
\newblock \showarticletitle{VRRobot: Robot actuated props in an infinite
  virtual environment}. In \bibinfo{booktitle}{\emph{2017 IEEE Virtual Reality
  (VR)}}. IEEE, \bibinfo{pages}{74--83}.
\newblock


\bibitem[\protect\citeauthoryear{Wang, Chen, Li, Cao, Luo, Zhang, Ou, Raiti,
  Yu, Patel, et~al\mbox{.}}{Wang et~al\mbox{.}}{2020}]%
        {wang2020movevr}
\bibfield{author}{\bibinfo{person}{Yuntao Wang}, \bibinfo{person}{Zichao Chen},
  \bibinfo{person}{Hanchuan Li}, \bibinfo{person}{Zhengyi Cao},
  \bibinfo{person}{Huiyi Luo}, \bibinfo{person}{Tengxiang Zhang},
  \bibinfo{person}{Ke Ou}, \bibinfo{person}{John Raiti}, \bibinfo{person}{Chun
  Yu}, \bibinfo{person}{Shwetak Patel}, {et~al\mbox{.}}}
  \bibinfo{year}{2020}\natexlab{}.
\newblock \showarticletitle{Movevr: Enabling multiform force feedback in
  virtual reality using household cleaning robot}. In
  \bibinfo{booktitle}{\emph{Proceedings of the 2020 CHI Conference on Human
  Factors in Computing Systems}}. \bibinfo{pages}{1--12}.
\newblock


\bibitem[\protect\citeauthoryear{Whitmire, Benko, Holz, Ofek, and
  Sinclair}{Whitmire et~al\mbox{.}}{2018}]%
        {whitmire2018haptic}
\bibfield{author}{\bibinfo{person}{Eric Whitmire}, \bibinfo{person}{Hrvoje
  Benko}, \bibinfo{person}{Christian Holz}, \bibinfo{person}{Eyal Ofek}, {and}
  \bibinfo{person}{Mike Sinclair}.} \bibinfo{year}{2018}\natexlab{}.
\newblock \showarticletitle{Haptic revolver: Touch, shear, texture, and shape
  rendering on a reconfigurable virtual reality controller}. In
  \bibinfo{booktitle}{\emph{Proceedings of the 2018 CHI Conference on Human
  Factors in Computing Systems}}. ACM, \bibinfo{pages}{86}.
\newblock


\bibitem[\protect\citeauthoryear{Yixian, Takashima, Tang, Tanno, Fujita, and
  Kitamura}{Yixian et~al\mbox{.}}{2020}]%
        {yixian2020zoomwalls}
\bibfield{author}{\bibinfo{person}{Yan Yixian}, \bibinfo{person}{Kazuki
  Takashima}, \bibinfo{person}{Anthony Tang}, \bibinfo{person}{Takayuki Tanno},
  \bibinfo{person}{Kazuyuki Fujita}, {and} \bibinfo{person}{Yoshifumi
  Kitamura}.} \bibinfo{year}{2020}\natexlab{}.
\newblock \showarticletitle{ZoomWalls: Dynamic Walls that Simulate Haptic
  Infrastructure for Room-scale VR Worlds}. In
  \bibinfo{booktitle}{\emph{Proceedings of the 33rd Annual ACM Symposium on
  User Interface Software and Technology}}. ACM.
\newblock


\bibitem[\protect\citeauthoryear{Yokokohji, Muramori, Sato, and
  Yoshikawa}{Yokokohji et~al\mbox{.}}{2005}]%
        {yokokohji2005designing}
\bibfield{author}{\bibinfo{person}{Yasuyoshi Yokokohji},
  \bibinfo{person}{Nobuhiko Muramori}, \bibinfo{person}{Yuji Sato}, {and}
  \bibinfo{person}{Tsuneo Yoshikawa}.} \bibinfo{year}{2005}\natexlab{}.
\newblock \showarticletitle{Designing an encountered-type haptic display for
  multiple fingertip contacts based on the observation of human grasping
  behaviors}.
\newblock \bibinfo{journal}{\emph{The International Journal of Robotics
  Research}} \bibinfo{volume}{24}, \bibinfo{number}{9} (\bibinfo{year}{2005}),
  \bibinfo{pages}{717--729}.
\newblock


\bibitem[\protect\citeauthoryear{Zhao, Kim, Wang, Le~Goc, and Follmer}{Zhao
  et~al\mbox{.}}{2017}]%
        {zhao2017robotic}
\bibfield{author}{\bibinfo{person}{Yiwei Zhao}, \bibinfo{person}{Lawrence~H
  Kim}, \bibinfo{person}{Ye Wang}, \bibinfo{person}{Mathieu Le~Goc}, {and}
  \bibinfo{person}{Sean Follmer}.} \bibinfo{year}{2017}\natexlab{}.
\newblock \showarticletitle{Robotic assembly of haptic proxy objects for
  tangible interaction and virtual reality}. In
  \bibinfo{booktitle}{\emph{Proceedings of the 2017 ACM International
  Conference on Interactive Surfaces and Spaces}}. ACM,
  \bibinfo{pages}{82--91}.
\newblock


\bibitem[\protect\citeauthoryear{Zhu, Chen, Han, and Wu}{Zhu
  et~al\mbox{.}}{2019}]%
        {zhu2019haptwist}
\bibfield{author}{\bibinfo{person}{Kening Zhu}, \bibinfo{person}{Taizhou Chen},
  \bibinfo{person}{Feng Han}, {and} \bibinfo{person}{Yi-Shiun Wu}.}
  \bibinfo{year}{2019}\natexlab{}.
\newblock \showarticletitle{HapTwist: creating interactive haptic proxies in
  virtual reality using low-cost twistable artefacts}. In
  \bibinfo{booktitle}{\emph{Proceedings of the 2019 CHI Conference on Human
  Factors in Computing Systems}}. \bibinfo{pages}{1--13}.
\newblock


\end{thebibliography}

\end{document}